\documentclass[journal]{IEEEtran}
\usepackage[T1]{fontenc}
\usepackage{amsmath}
\usepackage{subfigure}
\usepackage{graphicx}
\usepackage{url}
\usepackage{algorithm}
\usepackage{algorithmic}
\usepackage{amssymb}
\usepackage{color}
\usepackage{multimedia}
\usepackage{setspace}
\usepackage{amsthm}
\usepackage{multirow}

\renewcommand{\mathbf}{\boldsymbol}

\renewcommand{\Re}{{\mathbb{R}}}

\newcommand{\I}{\mathcal{I}}

\def\ee{{\boldsymbol{e}}}
\def\aa{{\boldsymbol{a}}}
\def\bb{{\boldsymbol{b}}}

\def\cc{{\boldsymbol{c}}}

\def\one{{\boldsymbol{1}}}
\def\zero{{\boldsymbol{0}}}

\def\L{{\boldsymbol{L}}}

\def\uu{{\boldsymbol{u}}}
\def\vv{{\boldsymbol{v}}}
\def\yy{{\boldsymbol{y}}}
\def\xx{{\boldsymbol{x}}}
\def\ww{{\boldsymbol{w}}}
\def\zz{{\boldsymbol{z}}}
\def\uu{{\boldsymbol{u}}}

\def\X{{\mathcal{X}}}
\def\L{{\mathcal{L}}}

\def\subj{\mbox{subj. to }}

\newcommand{\soft}{\mathrm{soft}}

\newcommand{\lag}{\mathcal{L}}
\newcommand{\bt}{\boldsymbol{\theta}}

\theoremstyle{definition}

\begin{document}
%
\title{Fast $\ell_1$-Minimization Algorithms\\ For Robust Face Recognition
}
%
%
%

\author{Allen~Y.~Yang,~\IEEEmembership{Member,~IEEE,} Zihan~Zhou,~\IEEEmembership{Student Member,~IEEE,} Arvind~Ganesh,~\IEEEmembership{Member,~IEEE,} S.~Shankar~Sastry,~\IEEEmembership{Fellow,~IEEE,} and Yi~Ma,~\IEEEmembership{Senior Member,~IEEE}
\thanks{A.~Yang and S.~Sastry are with the Department of Electrical Engineering and Computer Sciences, University of California, Berkeley, USA. Z.~Zhou is with the Department of Electrical and Computer Engineering, University of Illinois at Urbana-Champaign, Urbana, USA. A.~Ganesh is with the IBM Research, Bangalore, India. Y. Ma is with the Visual Computing Group, Microsoft Research Asia, Beijing, China.
Corresponding author: Allen Yang, Cory Hall, University of California, Berkeley, CA 94720. Email: yang@eecs.berkeley.edu. Tel: 1-510-643-5798. Fax: 1-510-643-2356.
}
\thanks{The authors gratefully acknowledge support by ONR N00014-09-1-0230, NSF CCF 09-64215, NSF IIS 11-16012, and ARL MAST-CTA W911NF-08-2-0004. An preliminary version of the work was previously published in the Proceedings of the International Conference on Image Processing in 2010 \cite{YangA2010-ICIP}.}
}

%
%

\maketitle

\begin{abstract}
$\ell_1$-minimization refers to finding the minimum $\ell_1$-norm solution to an underdetermined linear system $\bb=A\xx$. Under certain conditions as described in compressive sensing theory, the minimum $\ell_1$-norm solution is also the sparsest solution. In this paper, our study addresses the speed and scalability of its algorithms. In particular, we focus on the numerical implementation of a sparsity-based classification framework in robust face recognition, where sparse representation is sought to recover human identities from very high-dimensional facial images that may be corrupted by illumination, facial disguise, and pose variation. Although the underlying numerical problem is a linear program, traditional algorithms are known to suffer poor scalability for large-scale applications. We investigate a new solution based on a classical convex optimization framework, known as Augmented Lagrangian Methods (ALM). The new convex solvers provide a viable solution to real-world, time-critical applications such as face recognition. We conduct extensive experiments to validate and compare the performance of the ALM algorithms against several popular $\ell_1$-minimization solvers, including interior-point method, Homotopy, FISTA, SESOP-PCD, approximate message passing (AMP) and TFOCS. To aid peer evaluation, the code for all the algorithms has been made publicly available.
\end{abstract}


%
\IEEEpeerreviewmaketitle

\section{Introduction}
\label{sec:introduction}
Compressive sensing (CS) has been one of the hot topics in the
signal processing and optimization communities in the past ten
years. In CS theory
\cite{CandesE2006-ICM,DonohoD2004,BrucksteinA2007,ChenS2001-SIAM},
it has been shown that the minimum $\ell_1$-norm solution to an
underdetermined system of linear equations is also the sparsest
possible solution under quite general conditions. More
specifically, assume an unknown signal
$\xx_0\in\Re^n$, a measurement vector $\bb\in\Re^m$ ($m < n$), and a full-rank matrix $A\in\Re^{m \times n}$ such that $\bb= A\xx_0$. Recovering $\xx_0$ given $A$
and $\bb$ constitutes a non-trivial linear inversion problem,
since the number of measurements in $\bb$ is smaller than the
number of unknowns in $\xx_0$. A conventional solution
to this problem is \emph{linear least squares}, which finds the
minimum $\ell_2$-norm solution (or the solution of least
energy) to this system. However, if $\xx_0$
is sufficiently sparse (i.e., most of its entries in the canonical coordinates are zero),
and the sensing matrix $A$ is
\emph{incoherent} with the basis under which $\xx_0$ is sparse
(i.e., the identity matrix for the canonical coordinates), then $\xx_0$
can be exactly recovered by computing the minimum $\ell_1$-norm
solution:
\begin{equation}
(P_1): \quad \min_\xx \, \|\xx\|_1\quad \subj\quad \bb= A\xx.
\label{eq:l1}
\end{equation}

In practice, $\bb$ often contains noise. In such cases, the equality constraint can be relaxed, resulting in the constrained \emph{basis pursuit denoising} (BPDN) problem:
\begin{equation}
(P_{1,2}): \quad \min_\xx\,\|\xx\|_1\quad \subj\quad \|\bb- A\xx\|_2 \le \epsilon,
\label{eq:l1-l2}
\end{equation}
where $\epsilon > 0$ is a pre-determined noise level. A variant of this problem is also well known as the unconstrained BPDN problem with a scalar weight $\lambda$:
\begin{equation}
(QP_{\lambda}): \quad\min_{\xx}\frac{1}{2}\|\bb - A\xx\|_2^2 + \lambda \|\xx\|_1.
\label{eq:Lagrangian}
\end{equation}
Theoretical analysis of the BPDN problem \cite{ChenS2001-SIAM,CandesE2006-CPAM} has shown that, although exact recovery of the the ground-truth signal $\xx_0$ is not possible with noise in many cases (e.g., when the observation is corrupted by random Gaussian noise), it can be well approximated by the solution of $(P_{1,2})$ or $(QP_{\lambda})$.

\begin{figure}[t]
\centering
\includegraphics[height=0.7in]{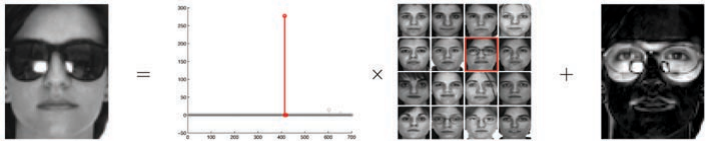}
\caption{Overview of the SRC framework for face recognition \cite{WrightJ2009-PAMI}. The test image is represented as a sparse linear combination of the training set as a whole, up to some sparse error due to corruption or occlusion. Note that the red coefficients correspond to training samples of the true subject.}
\label{fig:src}
\end{figure}

In this paper, we broadly refer to the above problems as $\ell_1$-minimization or $\ell_1$-min. The sparsity-seeking property of $\ell_1$-min optimization has been shown to have applications in many areas such as geophysics, speech recognition \cite{GemmekeHCB10}, image compression, processing, and enhancement \cite{BrucksteinA2007, YangJ2008-CVPR}, sensor networks \cite{BaronD2005, YangA2010-PIEEE}, and computer vision \cite{WrightJ2010-PIEEE}. Of all of these applications,  the \emph{sparse representation based classification} (SRC) framework proposed by \cite{WrightJ2009-PAMI} for face recognition is a representative and successful example. By casting the recognition problem as one of finding a sparse representation of the test image in terms of the training set as a whole, up to some sparse error, this framework has demonstrated striking recognition performance despite severe occlusion or corruption (see Figure~\ref{fig:src}) by solving a simple convex program. Later, it is shown in \cite{WagnerA2009-CVPR} that a local iterative process within the same framework can be employed to solve for an image transformation applied to the face region when the query image is misaligned, resulting in a state-of-the-art automatic face recognition system for access control scenarios.

While the $\ell_1$-min problems associated with CS can be formulated as a linear program (LP) and readily solved by classical methods in convex optimization, such as interior-point methods, the computational complexity of those classical methods is often too high for large-scale high-dimensional image data. In light of a large number of real applications in various fields, many new efficient algorithms have been proposed over the past decade. Meanwhile, to help the reader choose the best algorithm, several works exist in the literature which attempt to provide comprehensive reviews on the performance of $\ell_1$-min algorithms \cite{LorisI2009, BeckerS2009, Zibulevsky2010-SPM}, in addition to various amount of comparison experiments conducted in each individual paper that introduces new methods to $\ell_1$-min problems. 

A major limitation of existing works is that the performance of their algorithms is often benchmarked on \emph{synthetic data} only, and/or a couple of examples of simple signal and image processing applications with a \emph{single observation} (e.g., an image in image denoising). Meanwhile, we have seen that real-world data, especially high-resolution images, often demonstrate very special structures collectively. For instance, it is natural to expect that face images of different subjects are highly correlated with each other. Therefore, when applying existing $\ell_1$-min algorithms to complex problems, people often observe dramatically different behaviors compared to those reported in their original papers. For example, an algorithm called approximate message passing (AMP) is specifically designed for $\ell_1$-min problems when the dictionary is random Gaussian \cite{DonohoD2009-PNAS}. While outperforming many other methods on randomly generated synthetic data, it often fails to converge when applied to real face data. To this end, we do not believe there exists an optimal solution that would excel in every sparse optimization application.

\subsection{Contributions}
The goal of this paper to address the speed and scalability of $\ell_1$-min algorithms in the SRC framework for a real-world face recognition application. Our first contribution is a fast $\ell_1$-min solution based on a classical technique known as \emph{augmented Lagrangian methods} (ALM) \cite{Bertsekas1982}. Our solution is related to a previous solution known as the alternating direction methods (ADM) \cite{YangJ2009}. However, the discussion therein was restricted to the case when the dictionary is orthonormal or a randomly generated matrix. In this paper, we focus on the efficient implementation of ALM for face recognition applications.

Another contribution of the paper is a detailed comparison of the ALM algorithms with several state-of-the-art acceleration techniques for $\ell_1$-min problems, which include two classical solutions using \emph{interior-point method} and \emph{Homotopy method}, and several first-order methods including \emph{proximal-point methods} \cite{NesterovY1983-SMD,BeckA2009,BeckerS2009}, \emph{parallel coordinate descent} (PCD) \cite{Elad2007-ACHA}, \emph{approximate message passing}
(AMP) \cite{DonohoD2009-PNAS}, and \emph{templates for convex cone solvers} (TFOCS) \cite{BeckerS2010}. To set up the stage for a fair comparison and help the reader gain a basic understanding of the sparse optimization literature, we provide an extensive review of these techniques with an emphasis to their conceptual connections and their computational complexity in different sparse representation settings. 
\footnote{Due to the overwhelming volume of the sparse optimization literature, it is impossible to discuss and compare all the existing methods in a single paper. Methods that are not discussed in this paper include GPSR \cite{FigueiredoM2007}, SpaRSA \cite{WrightS2008}, SPGL1 \cite{BergFriedlander2008}, NESTA \cite{BeckerS2009}, SALSA \cite{Afonso2010-TIP}, GLMNET \cite{FriedmanJ2010}, and \emph{Bregman iterative algorithm} \cite{YinW2008}, just to name a few. Nevertheless, vast majority of the existing algorithms are variants of those benchmarked in this paper, and share many common properties with them. We will elaborate on their connections more later. Also, in the literature, there exist greedy algorithms to estimate sparse signals, such as \emph{Orthogonal Matching Pursuit} (OMP) \cite{DavisG1997} and its many variants and extensions \cite{BrucksteinA2007, NeedellD2008, DaiW2009}. OMP was originally proposed to solve a related optimization problem called $\ell_0$-minimization. In this paper, we are not concerned about these pursuit-type greedy algorithms, but instead refer the reader to \cite{TroppJ2004,BrucksteinA2007,TroppJ2010} for a more detailed treatment of the greedy approach.}

To concretely demonstrate the performance of ALM and the other algorithms, we have compiled a thorough benchmark using both synthetic data and real high-dimensional image data in face recognition. The ALM algorithms compare favorably among a wide range of state-of-the-art $\ell_1$-min algorithms, and more importantly are very suitable for large-scale face recognition and alignment problems in practice. To aid peer evaluation, all algorithms discussed in this paper have been made available on our website as a MATLAB toolbox: \url{http://www.eecs.berkeley.edu/~yang/software/l1benchmark/}.

Finally, we need to point out that there have been several recent studies in the community, which seek alternative approaches to robust face recognition where solving the nonsmooth sparse optimization problems such as $\ell_1$-min could be totally mitigated. We refer the interested reader to discussions in \cite{ZhangL2011-ICCV,WrightJ2011-arXiv}. In a nutshell, most of these alternative solutions achieve faster speed by some tradeoffs that sacrifice the recognition accuracy, especially when the data could contain high levels of data noise, corruption, and/or spatial misalignment. In contrast, the main focus of this paper is accelerated convex optimization techniques that provably converge to the global optimum of the $\ell_1$-min objective function \eqref{eq:l1} and, more importantly, without sacrificing the recognition accuracy. 

\section{A Review of Robust Face Recognition via Sparse Representation}
\label{sec:face-rec}

Since the focus of this paper is on the efficiency of various $\ell_1$-min methods for sparse representation based face recognition, we begin our discussion with a brief review of the related face recognition techniques. In this paper, all data are assumed in the real domain. The concatenation of two vectors will be written following the MATLAB convention: $[\xx_1; \xx_2]\doteq \left[\begin{smallmatrix}\xx_1\\ \xx_2 \end{smallmatrix}\right]$;  $[\xx_1, \xx_2]\doteq \left[\begin{smallmatrix}\xx_1 & \xx_2 \end{smallmatrix}\right]$. We denote by $\one$ a vector whose components are all one with dimension defined within the context. We represent the Euclidean or $\ell_2$-norm by $\|\cdot\|_2$ and the $\ell_1$-norm by $\|\cdot\|_1$. The notation $\|\cdot \|$ represents the $\ell_2$-norm for vectors and the spectral norm for matrices.

\subsection{Dense error correction via $\ell_1$-minimization}

In the face recognition literature, it is known that a well-aligned frontal face image $\bb\in\Re^m$ of a human subject under different illuminations lies closely to a low-dimensional subspace, called face subspace \cite{BelhumeurP1997-PAMI,BasriR2003-PAMI}. Therefore, given a known subspace class $i$ and sufficient training samples,  $A_i = [\vv_{i,1}, \vv_{i,2}, \cdots, \vv_{i,n_i}] \in \Re^{m\times n_i}$, where $\vv_{i,j}$ represents the $j$-th training image from the $i$-th subject stacked in the vector form, $\bb$ from the $i$-th class can be represented as $\bb = A_i\xx_i$. 

Now given $C$ subjects, the SRC framework proposed in \cite{WrightJ2009-PAMI} tries to determine the identity of the query image $\bb$ by seeking the sparsest linear representation of $\bb$ with respect to all the training examples:
\begin{equation}
\bb = [A_1, A_2, \cdots, A_C] [\xx_1; \xx_2; \cdots; \xx_C] \doteq A\xx.
\label{eq:intro-classification}
\end{equation}

Clearly, if $\bb$ is a valid test image, it must lie in one of the $C$ face subspaces.
Therefore, the corresponding representation in \eqref{eq:intro-classification} has a sparse representation $\xx = [\cdots; \zero; \xx_i; \zero; \cdots ]$: on average only a fraction of $\frac{1}{C}$ coefficients are nonzero, and the dominant nonzero coefficients in sparse representation $\xx$ reveal the true subject class.\footnote{See Section~\ref{sec:experiment} and \cite{WrightJ2009-PAMI} for more details about the implementation of the classifier.}
\begin{figure}[t]
\centering
\includegraphics[height=1.5in]{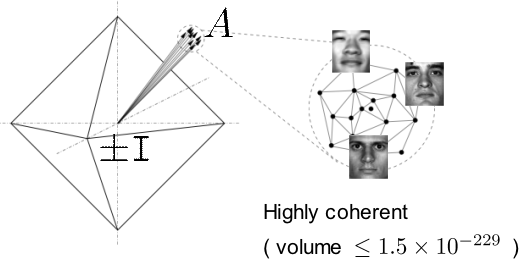}
\caption{The CAB model for face recognition. The raw images of human faces expressed as columns of $A$ are clustered with very small variance (Courtesy of John Wright \cite{WrightJ2008-IT}).}
\label{fig:cross-and-bouquet}
\end{figure}

In addition to the possible illumination changes, $\bb$ is often occluded or corrupted in practice. In \cite{WrightJ2009-PAMI}, a modified sparse representation model was proposed as $\bb = A\xx + \ee$, where $\ee\in\Re^m$ is another unknown vector whose nonzero entries correspond to the corrupted pixels. Consequently,  let $\ww\doteq[\xx; \ee]$, and it can be estimated jointly by $\ell_1$-min:
\begin{equation}
\min\|\ww \|_1 \quad \subj \quad \bb = [A,\ I]\ww = A\xx + \ee.
\label{eq:CAB}
\end{equation}
Here, a key observation is that the new dictionary $[A,\ I]$ has very special structures. It was dubbed \emph{cross-and-bouquet} (CAB) model in \cite{WrightJ2008-IT} in the following sense: The columns of A are highly correlated, as the convex hull spanned by all face images of all subjects occupies an extremely tiny portion of the ambient space. These vectors are tightly bundled together as a ``bouquet,'' whereas the vectors associated with the identity matrix and its negative $\pm I$ form a standard ``cross'' in $\Re^{m}$, as shown in Figure \ref{fig:cross-and-bouquet}.

The implication of this special model for face recognition is at least two-fold. On the theory side, it enables exactly recovery of both $\xx$ and $\ee$ via solving the $\ell_1$-min problem \eqref{eq:CAB} even when $\ee$ is dense (i.e., up to nearly 100\% of the pixels are corrupted), as long as the bouquet is sufficiently tight and the dimensions of the signal $\xx$ and the observation $\bb$ are sufficiently high \cite{WrightJ2008-IT}.

On the practice side, however, it poses new challenges to existing $\ell_1$-min algorithms as the dictionary $A$ is ill-conditioned, or more specifically, highly coherent in CS jargon. Meanwhile, in real-world applications, it is important for a face recognition system to efficiently handle thousands or even more subjects while the dimension of each image remains roughly unchanged. Thus, a preferred algorithm should scale well in terms of $C$ and the total number of images $n$.

\subsection{Face Alignment} 
Another important factor that affects face recognition is image misalignment, which is often caused by an inaccurate face detector applied to images collected in uncontrolled environments (see Figure \ref{fig:alignment-expt} for example). Clearly, when a query image is not aligned well with the training images, the face subspace model in \eqref{eq:intro-classification} and \eqref{eq:CAB} will not be satisfied. Recently, \cite{WagnerA2009-CVPR} shows that this problem can be solved nicely within the sparse representation framework by iteratively optimizing a series of linear approximate problems that minimize the sparse registration error $\ee$ in \eqref{eq:CAB} while
the query image $\bb$ is under an image transformation.

More specifically, suppose the ground truth image $\bb_0$ is subject to some
misalignment caused by a transformation $\tau\in T$, where $T$ is a finite-dimensional
group of transformations acting on the image domain. As a result, we observe
the warped image $\bb = \bb_0 \circ \tau^{-1}$. This relationship can be 
rewritten in the CAB model as: $\bb \circ \tau = A \xx + \ee$.
Naturally, we would like to use the sparsity as a cue
for finding the correct deformation $\tau$, such as solving the following optimization problem:
\begin{equation}
\min_{\xx,\ee,\tau\in T}\|\xx\|_1 + \|\ee\|_1 \quad \subj \quad \bb \circ \tau = A \xx + \ee.
\label{eq:global-alignment}
\end{equation}

However, this is a difficult nonconvex optimization problem. Due to the concern of local minima, directly solving \eqref{eq:global-alignment} may simultaneously align the query
image to different subjects in the database. Therefore, it is more
appropriate to seek the best alignment with respect to each subject $i$ in
the database \cite{WagnerA2009-CVPR}:
\begin{equation}
\hat{\tau_i} = \arg \min_{\xx,\ee,\tau_i\in T}\|\ee\|_1 \quad \subj \quad \bb \circ \tau_i = A_i \xx + \ee.
\label{eqn-align}
\end{equation}

In \eqref{eqn-align}, $\|\xx\|_1$ is not penalized, since $A_i\in \Re^{m\times
n_i}$ only contains the images of subject $i$ and $\xx$ is no
longer expected to be sparse. While \eqref{eqn-align} is still
nonconvex, when a good initial estimation for the
transformation is available, e.g., from the output of a face
detector, one can refine this initialization to an estimate of
the true transformation by linearizing about the the
current estimate of $\tau_i$, which leads to a convex problem:
\begin{equation}
\min_{\xx,\ee,\Delta \tau_i\in T}\|\ee\|_1 \quad \subj \quad \bb\circ \tau_i + J_i\Delta \tau_i = A_i \xx + \ee.
\label{eqn-align-linear}
\end{equation}
Here, $J_i = \frac{\partial}{\partial \tau_i}(\bb \circ \tau_i) \in
\Re^{m\times q_i}$ is the Jacobian of $\bb \circ \tau_i$ with
respect to the transformation parameters $\tau_i$, and $\Delta
\tau_i$ is the current update step with respect to (w.r.t.) $\tau_i$.

During each iteration $j$, the current alignment parameters $\tau_i^j$ correct the observation as
$\bb_i^j = \bb\circ \tau_i^j$. Denote $B_i^j = [A_i, -J_i^j]\in
\Re^{m\times (n_i+q_i)}$ and $\ww = [\xx^T, \Delta\tau_i^T]^T$, then the update $\Delta\tau_i$ can be computed by solving the problem:
\begin{equation}
\min_{\ww,\ee}\|\ee\|_1 \quad \subj \quad \bb_i^j = B_i^j\ww+\ee.
\label{eqn-align-simple}
\end{equation}

The interested reader is referred to \cite{WagnerA2009-CVPR} for more details about the effectiveness of this approach. In this paper, we will focus on the fast solutions to \eqref{eqn-align-simple}, and the face alignment problem.

\section{Classical methods for $\ell_1$-min problems}

In this section, we lay the foundation for our discussion on $\ell_1$-min algorithms by reviewing two classical methods, the interior-point method and the Homotopy method. The two methods will be used extensively in Section \ref{sec:experiment} to provide baseline performance and estimate ground-true sparse signals in the experiment.

\label{sec:classical}
\subsection{Primal-dual interior-point algorithm}
We first consider a classical approach as the baseline to solving $\ell_1$-min, called \emph{primal-dual interior-point algorithm} (PDIPA). The PDIPA framework is usually attributed to the works of \cite{FrischK1955,KarmarkarN1984,MegiddoN1989,MonteiroR1989-I,KojimaM1993}. For the sake of simplicity, here we assume that the sparse solution $\xx$ is nonnegative.\footnote{This constraint can be easily removed by considering the linear system $\bb = [A, -A][\xx_+; \xx_-]$, where $\xx_+$ and $\xx_-$ are the vectors that collect the positive and negative coefficients of $\xx$, respectively: $\xx = \xx_+ - \xx_-, \xx_+\ge 0, \xx_-\ge 0.$
} 

Under this assumption, $(P_1)$ can be rewritten as a LP:
\begin{equation}
\begin{array}{rllrl}
 & \mbox{\bf Primal (P)}& & & \mbox{\bf Dual (D)} \\
\min_\xx& \one^T\xx; & & \max_{\yy,\zz} & \bb^T\yy.\\
\subj & A\xx = \bb & & \subj & A^T\yy + \zz = \one\\
&  \xx \ge 0 & & & \zz\ge 0
\end{array}
\label{eq:linear-program}
\end{equation}

The basic idea of primal-dual inter-point method is to iteratively formulate the inequality constrained problem \eqref{eq:linear-program} as an equality constrained problem, which can be solved by Newton's method, using the barrier method \cite{FrischK1955, BoydS2004}. Hence, the complexity of PDIPA is dominated by the Newton update step, which is bounded by $O(n^3)$. As we will see later from our experiment results (e.g., Figure~\ref{fig:recognition-k}), PDIPA is among the ones which are most sensitive to the size of the problem, hence is not suitable for large-scale real-world applications.

Since solving the Newton system exactly is computational expensive for large $\ell_1$-min problems, fast methods that approximate its solution have been exploited in the literature. In particular, \cite{KimS2007} uses an iterative method, namely, preconditioned conjugate gradient (PCG) \cite{KelleyC1995,NocedalJ2006}, to approximately solve the Newton system and develops an interior-point method that solves $(QP_{\lambda})$ \eqref{eq:Lagrangian}. This overall algorithm is called \emph{truncated Newton interior-point method} (TNIPM).\footnote{An implementation of TNIPM called L1LS is available at \url{http://www.stanford.edu/~boyd/l1_ls/}.} By carefully choosing the preconditioner, it is comparable to first-order methods in solving large problems with modest accuracy, while retaining the ability of second-order method to solve them with high accuracy at relatively small additional computational cost.

Note that since TNIPM/L1LS is designed for $(QP_{\lambda})$, to recover the exact  solution for $(P_1)$, it is necessary for $\lambda$ to gradually decrease to zero. This is also the case for another two algorithms benchmarked in this paper, namely FISTA and SESOP-PCD. In Section~\ref{sec:experiment}, we will revisit this issue in more details by comparing algorithms that solve $(P_1)$ directly with those solving the relaxed problem $(QP_{\lambda})$.

\subsection{Homotopy methods}

Homotopy methods in sparse optimization are specifically designed to take advantage of the properties of $\ell_1$-min. The approach was first studied in the context of LASSO \cite{OsborneM2000}, which inspired a solution to the \emph{forward stagewise linear regression} problem called LARS \cite{EfronB2004} and eventually led to the Homotopy algorithms for basis pursuit in \cite{MalioutovD2005,DonohoD2006}.\footnote{Homotopy package \cite{AsifM2008}: \url{http://users.ece.gatech.edu/~sasif/Homotopy/}.}

The fundamental idea of Homotopy is the following: In solving the noisy version of basis pursuit $(QP_{\lambda})$, the method exploits the fact that the objective function $F(\xx)$ undergoes a homotopy from the $\ell_2$ constraint to the $\ell_1$ objective as $\lambda$ decreases. More specifically, when $\lambda\rightarrow \infty$, $\xx^*_\lambda=0$; when $\lambda\rightarrow 0$, $\xx^*_\lambda$ converges to the solution of $(P_1)$. Furthermore, one can show that the solution path $\X\doteq \{\xx^*_\lambda:  \lambda \in [0, \infty)\}$ is piecewise constant as a function of $\lambda$ \cite{EfronB2004}. Therefore, in constructing a decreasing sequence of $\lambda$, it is only necessary to identify those ``breakpoints'' that lead to changes of the support set of $\xx^*_\lambda$,  namely, either a new nonzero coefficient added or a previous nonzero coefficient removed.

If one were to directly compute the gradient of $F$ in \eqref{eq:Lagrangian}, one obstacle is that the $\ell_1$-norm term $g(\xx)$ is not globally differentiable. Therefore, we consider the \emph{subdifferential} of $\|\xx\|_1$ defined as follows:
\begin{equation}
\uu(\xx) \doteq \partial\|\xx\|_1 = \left\{\uu\in\Re^n : \begin{matrix}u_i=\mbox{sgn}(x_i), x_i\not=0\\ u_i\in [-1, 1], x_i=0 \end{matrix}\right\}.
\label{eq:l1-subdifferential}
\end{equation}
The Homotopy algorithm operates in an iterative fashion with an initial value $\xx^{(0)}=0$. In each iteration w.r.t. a nonzero $\lambda$, the condition $0\in \partial F(\xx)$ leads to:
\begin{equation}
\cc(\xx)=A^T\bb - A^TA\xx \in \lambda \uu(\xx).
\label{eq:Homotopy-condition}
\end{equation}
Hence, according to the definition \eqref{eq:l1-subdifferential}, we maintain a sparse support set $\I \doteq \{i: |\cc^{(k)}_i(\xx)|=\lambda \}$ at the $k$-th iteration. Then the algorithm computes the update direction and stepsize only for the nonzero coefficients of $\xx^{(k)}$ identified by $\I$. In summary, since the Newton update only involves nonzero coefficients in $\I$, which could be a very small number when $\xx$ is sparse, the computational cost of the Homotopy algorithm for $\ell_1$-min is bounded by $O(dm^2 + dmn)$ if it correctly recovers a $d$-sparse signal in $d$ steps, a significant improvement from the interior-point methods \cite{DonohoD2006}. It is also clear from the equation that when the sparsity $d$ and the observation dimension $m$ grow proportionally with the signal dimension $n$, the worst-case complexity is still bounded by $O(n^3)$, a major drawback of Homotopy methods especially for recovering non-sparse signals.

Therefore, in the next section, we will turn to another category of fast $\ell_1$-min algorithms, known as \emph{first-order methods}. These algorithms enjoy much better worst-case complexity than interior-point and Homotopy methods, hence scale well for large-scale problems such as face recognition.

\section{First-Order Methods}
\label{sec:first_order_methods}

In optimization, {\it first-order methods} refer to those algorithms that have at most linear local error, typically based on local linear approximation. In the context of $\ell_1$-min, first-order methods differ from the previous classical approaches in that they explicitly make use of the structure of the subdifferential of the $\|\cdot\|_1$. The advantage of first-order methods is that the computational complexity per iteration is greatly reduced, albeit at the expense of increasing the number of iterations as compared to the interior-point methods. Here we consider four most visible algorithms in recent years, namely,
\emph{proximal-point methods} \cite{NesterovY1983-SMD,BeckA2009,BeckerS2009}, \emph{parallel coordinate descent} (PCD) \cite{Elad2007-ACHA}, \emph{approximate message passing}
(AMP) \cite{DonohoD2009-PNAS}, and \emph{templates for convex cone solvers} (TFOCS) \cite{BeckerS2010}.

Before proceeding, we first introduce the proximal operator of a convex function $g$ of $\xx\in \Re^n$, which is defined as
\begin{equation}
\textup{prox}_g(\xx) \doteq \arg\min_{\uu} g(\uu) + \frac{1}{2}\|\uu - \xx\|_2^2.
\end{equation}
It is well known that for $\ell_1$-min problems where $g(\xx) = \alpha\|\xx\|_1$, the proximal operator has a closed-form expression called the soft-thresholding or shrinkage operator, $\soft(\xx, \alpha)$, which is defined element-wise as follows \cite{CombettesP2005}:
\begin{equation*}
\soft(\xx, \alpha)_i = \mathrm{sign}(x_i)\cdot \max\{|x_i|-\alpha, 0\}, \quad i = 1, 2, \ldots, n.
\end{equation*}

The implementation of first-order algorithms mainly involves elementary linear algebraic operations such as vector addition, matrix-vector multiplication, and soft-thresholding. These operations are much cheaper computationally compared to matrix inversion and matrix factorization that are commonly required in other conventional methods.

\subsection{Proximal-Point Methods}

Recall the following objective function in Section \ref{sec:introduction}:
\begin{equation}
F(\xx) = \frac{1}{2} \, \|\bb-A\xx\|_2^2 + \lambda \|\xx\|_1 \doteq f(\xx) + \lambda g(\xx).
\label{eqn:lagrange_ist}
\end{equation}
Note that $F(\xx)$ to be minimized is a composite of two
functions with very different properties. On one hand,
$f(\cdot)$ is a smooth, convex function with a Lipschitz
continuous gradient given by $\nabla f(\xx) = A^T (A\xx -
\bb)$. The associated Lipschitz constant $L_f$ of $\nabla f(\cdot)$ is given by the
spectral norm of $A^T A$, denoted by $\|A^T A\|_2$. On the
other hand, $g(\cdot)$ is a continuous, convex but non-smooth function. 

In general, proximal-point methods work by generating a sequence of iterates $\{\xx_k, k = 0, 1, \ldots\}$, and at each iteration solving the following subproblem which approximates $F(\xx)$:
\begin{eqnarray}
\xx_{k+1} & = & \arg\min_{\xx}\{f(\xx_k) + (\xx-\xx_k)^T \nabla f(\xx_k) + \nonumber \\
& & \frac{\alpha_k}{2} \, \|\xx-\xx_k\|_2^2 + \lambda g(\xx)\},
\label{eq:ISTA}
\end{eqnarray}
for some $\alpha_k >0$. Using the soft-thresholding operator, the above subproblem has a closed-form solution:
\begin{equation}
\xx_{k+1} = \soft\left(\xx_k - \frac{1}{\alpha_k} \nabla f(\xx_k),\frac{\lambda}{\alpha_k}\right).
\end{equation}

Obviously, the convergence behavior of the above scheme depends on the choice of $\alpha_k$. For example, the popular \emph{iterative soft-thresholding algorithm} (ISTA) \cite{CombettesP2005, DaubechiesI2004, HaleE2007, WrightS2008} employs a fixed choice of $\alpha_k$ related to $L_f$. In \cite{BeckA2009}, assuming $\alpha_k = L_f$, one can show that ISTA has a sublinear convergence rate that is no worse than $O(1/k)$:
\begin{equation}
F(\xx_k) - F(\xx^*) \le \frac{L_f\|\xx_0 - \xx^* \|^2}{2k}, \quad \forall \, k.
\end{equation}
Meanwhile, an alternative way of determining $\alpha_k$ at each iteration is used in SpaRSA \cite{WrightS2008}, which is based on the \emph{Barzilai-Borwein} equation \cite{BarzilaiJ1988}. It has been shown that SpaRSA has the same convergence rate of  $O(1/k)$ for $\ell_1$-min problems \cite{HagerPZ11}.

While the above methods enjoy a much lower computation complexity per iteration, in practice people have observed that it converges quite slowly in terms of the number of iterations. Recently, \cite{BeckA2009} proposes a \emph{fast iterative soft-thresholding algorithm} (FISTA), which has a significantly better convergence rate. The key idea behind FISTA is that, instead of forming a quadratic approximation of $F(\xx)$ at $\xx_k$ at the $k$-th iteration as in \eqref{eq:ISTA}, it uses a more carefully chosen sequence $\yy_k$ for that purpose. This is known as the {\it ravine} step as described in \cite{NesterovY1983-SMD} and leads to the following FISTA iterations:
\begin{equation}
\left \{
\begin{array}{rcl}
\xx_{k} & = & \soft\left(\yy_k - \frac{1}{L_f} \nabla f(\yy_k),\frac{\lambda}{L_f}\right), \\
t_{k+1} & = & \frac{1+\sqrt{1+4t_k^2}}{2}, \\
\yy_{k+1} & = & \xx_k + \frac{t_k - 1}{t_{k+1}} (\xx_k - \xx_{k-1}),
\end{array}
\right .
\label{eqn:fista_iteration}
\end{equation}
where $\yy_1 = \xx_0$ and $t_1 = 1$. The same idea has also been applied to solve the constrained problem $(P_{1,2})$ in \cite{BeckerS2009}, yielding the so-called \emph{Nesterov's algorithm} (NESTA).\footnote{NESTA package: \url{http://www-stat.stanford.edu/~candes/nesta/}.}  Both algorithms enjoy the same non-asymptotic convergence rate of $O(1/k^2)$ in the $\ell_1$-min setting. 
The interested reader may refer to \cite{BeckA2009} for a proof of the above result, which extends the original algorithm of Nesterov \cite{NesterovY2007} devised only for smooth functions that are everywhere Lipschitz continuous.

\subsection{Parallel Coordinate Descent Algorithm} An alternative way to use the soft-thresholding
operator is considered in \cite{Elad2007-ACHA}. It starts with
the observation that if in each iteration we update $\xx$ one
entry at a time, which is known as the coordinate descent (CD) in the literature, then each of these updates can be
obtained in closed form. More precisely, for updating the
$i$-th entry of the current estimate $\xx_k$, one needs to solve
the following problem:
\begin{equation}
g(v) = \frac{1}{2}\|\bb-A\xx_k -\aa_i(v-\xx_k[i]) \|_2^2 + \lambda |v|,
\end{equation}
where $\aa_i$ is the $i$-th column of $A$. The optimal solution of $v$ is again given by the soft-thresholding operator:
\begin{equation}
v^* = \soft \left(\frac{\aa_i^T(\bb-A\xx_k)}{\|\aa_i\|_2^2} + \xx_k[i], \frac{\lambda}{\|\aa_i\|_2^2}\right).
\end{equation}

As pointed out in \cite{Elad2007-ACHA}, while a sequence of
such rounds of $n$ updates (addressing each coordinates of
$\xx$ in certain order) is necessarily converging, it requires
explicit access to each column of $A$. This is however
computational inefficient because many transformations
associated with $A$, such as wavelet transform, are actually
computed via a fast recursive scheme, rather than direct
matrix-vector multiplication. For this reason,
\cite{Elad2007-ACHA} proposes to merge such descent steps into
a joint step using a simple addition, leading to the following \emph{parallel} update rule:
\begin{eqnarray}
\vv^* & = & \sum_{i=1}^n \ee_i \cdot \soft \left(\frac{\aa_i^T(\bb-A\xx_k)}{\|\aa_i\|_2^2} + \xx_k[i], \frac{\lambda}{\|\aa_i\|_2^2}\right) \nonumber \\
& = & \soft\left( WA^T(\bb-A\xx_k)+\xx_k, W\lambda\right),
\end{eqnarray}
where $W=\textup{diag}(A^TA)^{-1}$. This updating rule is
further combined with a line search in \cite{Elad2007-ACHA} to
ensure that the direction is indeed descending, resulting in a
new iterative algorithm of the form:
\begin{equation}
\xx_{k+1} =  \xx_k + \beta(\vv^*-\xx_k),
\end{equation}
which is referred as the PCD algorithm. 

In \cite{Elad2007-ACHA}, PCD is further accelerated using a \emph{sequential subspace optimization} (SESOP) technique. The key idea of SESOP is that instead of searching along a single direction $\vv^*-\xx_k$ as in the PCD algorithm, a set of directions of the last $M$ propagation steps is also included, where $M$ is specified by the user. The interested reader may refer to \cite{Zibulevsky2010-SPM} for a more detailed comparison between the proximal-point methods and the PCD algorithm under various synthetic and image processing settings.\footnote{A MATLAB implementation of both SESOP-PCD and FISTA is available at \url{http://iew3.technion.ac.il/~mcib/sesop.html}.} 



\subsection{Approximate Message Passing}
More recently, Donoho et al.
\cite{DonohoD2009-PNAS} have shown that iterative
soft-thresholding can be understood as an approximate solution to $(P_1)$ via a belief propagation framework
\cite{BaronD2010}. In this graph-theoretic framework, the
problem of basis pursuit is modeled by a \emph{factor graph}
$G=\{X, F, E\}$, which is a complete bipartite graph with
variable nodes $X = \{x_1, x_2, \cdots, x_n\}$, factor nodes $F
= \{f_1, f_2, \cdots, f_m\}$, and edges $E=X\times F = \{(x_i,
f_j): x_i\in X, f_j\in F \}$. Assuming the probability
distribution of each variable $x_i$ satisfies a Laplace prior
$\frac{1}{C} \exp(-\beta | x_i |)$ and each factor node $f_j$
is a Dirac delta function $\delta(b_j = (A\xx)_j)$, then the
overall joint probability for an observation vector $\bb$ and
solution $\xx$ is the following function:
\begin{equation}
p(\bb, \xx) = \frac{1}{Z}\prod_{i=1}^{n} \exp(-\beta | x_i |) \prod_{j=1}^m \delta(b_j = (A\xx)_j).
\end{equation}
In a sense, the above factor graph describes a decoding process, where each unknown variable $x_i$ is assumed a prior distribution and each ``parity-check'' function $f_j$ ensures the recovered code satisfies the linear constraint $\bb = A\xx$.
Furthermore, when $\beta \rightarrow \infty$, the joint probability $p(\bb, \xx)$ will concentrate around the sparse solution of $\ell_1$-min \eqref{eq:l1}. Hence, the $\ell_1$-min solution can be estimated by iteratively computing the marginal distribution $p(x_i)$ for each given variable $x_i$.

In Bayesian networks, the marginal distribution of a factor graph $G$ can be estimated by standard message-passing algorithms \cite{KschischangF2001-TIT}. However, the exact message-passing algorithm applied to the complete bipartite graph $G$ is not cost effective, as the graphical model is dense. Furthermore, the convergence of the algorithm to the optimal basis pursuit solution cannot be guaranteed, as graph $G$ represents a loopy network.\footnote{In \cite{BaronD2010}, the drawback of message passing is mitigated by imposing a sparse dictionary $A$, which reduces both the number of loops and the cost of belief propagation.} To address these issues, an \emph{approximate message-passing} (AMP) algorithm was derived in \cite{DonohoD2009-PNAS}.\footnote{AMP package: \url{http://bigwww.epfl.ch/kamilov}.} Specifically, if the variables $\xx$ satisfy the Laplace prior and the dictionary $A$ is a Gaussian random matrix, when $m, n\rightarrow \infty$ proportionally, the update rule for $\xx$ in $G$ is approximated by the following equations:
\begin{equation*}
\left \{
\begin{array}{rcl}
\xx_{k+1} &=& \soft(A^T\zz_{k} + \xx_{k}, \tau_{k}),\\
\zz_{k+1} &=& \bb - A\xx_{k+1} + \frac{1}{\delta}\zz_k \langle \frac{\partial}{\partial \xx} \soft(A^T\zz_k + \xx_{k}, \tau_k) \rangle,\\
\tau_{k+1} &=& \frac{\tau_k}{\delta} \langle \frac{\partial}{\partial \xx} \soft(A^T\zz_{k} + \xx_{k}, \tau_k) \rangle,
\end{array}
\right.
\end{equation*}
where $\langle \cdot \rangle$ denotes the average of a vector.

As shown in both \cite{DonohoD2009-PNAS} and our experiment results in this paper, this method achieves the state-of-the-art performance in solving $(P_1)$ \emph{when $A$ is a random Gaussian matrix}. However, it cannot handle problems in which $A$ violates this assumption, which is indeed the case for the face recognition applications. 

\subsection{Templates for Convex Cone Solvers (TFOCS)}

TFOCS is a relatively new framework proposed in \cite{BeckerS2010} which proposes to use the soft-thresholding operator to solve the dual problems of $\ell_1$-min.\footnote{TFOCS package: \url{http://tfocs.stanford.edu/download/}.} It considers a class of algorithms that deal with constrained convex optimization problems of the following type:
\begin{equation}
\min_\xx \, f(\xx) \quad \subj \quad \mathcal{A}(\xx) + \bb \in \mathcal{K},
\end{equation}
where $f(\cdot)$ is a convex function, $\mathcal{A}(\cdot)$ is a linear operator, $\bb$ is a fixed point, and $\mathcal{K}$ is a closed, convex cone. One can see that both $(P_1)$ and $(P_{1,2})$ fall in this category.

In a nutshell, the main idea behind TFOCS is to solve the dual problem by a generalized projected gradient ascent technique, and in the process obtain the primal optimal solution as well. However, for most sparse recovery problems, the dual cost function is not smooth. To overcome this issue, \cite{BeckerS2010} recommended adding a smoothing term to the primal cost function. In the context of $\ell_1$-min, the problem $(P_1)$ would reduce to the following form:
\begin{equation}
\min_\xx \, \|\xx\|_1 + \mu \, \phi(\xx) \quad \subj \quad A\xx - \bb = \mathbf{0},
\label{eq:TFOCS-l1-min}
\end{equation}
where $\mu > 0$ is a smoothing parameter and $\phi(\cdot)$ is a strongly convex function satisfying
\begin{equation}
\phi(\xx) \geq \phi(\xx_0) + \frac{1}{2}\|\xx-\xx_0\|_2^2,
\end{equation}
for some fixed point $\xx_0$. For instance, if $\phi(\cdot)$ is chosen as $\phi(\xx) = \frac{1}{2}\|\xx - \xx_0\|_2^2$, with the choice of $\xx_0$ specified later, then the conic Lagrangian of \eqref{eq:TFOCS-l1-min} is given by
\begin{equation}
\L_\mu(\xx,\bt) = \|\xx\|_1 + \frac{\mu}{2}\|\xx- \xx_0\|_2^2 - \bt^T (A\xx - \bb),
\end{equation}
where $\bt$ is a vector of Lagrange multipliers.

By definition, let $\L_\mu(\bt) \doteq \min_\xx \, \L_\mu(\xx, \bt)$, then the dual problem of \eqref{eq:TFOCS-l1-min} is given by $\max_{\bt} \, \L_\mu (\bt)$. Since $\L_\mu(\xx,\bt)$ is strongly convex in $\xx$, a unique minimizer $\hat{\xx}_\mu(\bt)$ exists, which is given by the soft-thresholding operator: 
\begin{equation}
\hat{\xx}_\mu(\bt) = \soft\left(\xx_0 + \frac{1}{\mu}A^T\bt ,\frac{1}{\mu}\right).
\end{equation}
Furthermore, $\L_\mu(\bt)$ is a smooth, concave function whose gradient is given by $\nabla \L_\mu(\bt) = \bb - A \hat{\xx}_\mu(\bt)$. Therefore, the following iterative scheme can be constructed to update the primal and dual variables based on the \emph{first-order projected gradient methods} proposed in \cite{Tseng2008,AuslenderJ2006-SJO}:
\begin{equation}
\left \{
\begin{array}{rcl}
\xx_k & = & \soft\left(\xx_0 + \frac{1}{\mu}A^T\bt_k ,\frac{1}{\mu}\right), \\
\bt_{k+1} & = & \bt_k + t_k  (\bb - A\xx_k),
\end{array}
\right .
\end{equation}
where $\{t_k\}$ is a sequence of step sizes satisfying $t_k \leq \mu/ \|A^T A\|$ for all $k$. Here we note that it has a nice property that, for sufficiently small $\mu$, the solution obtained by the above iterative scheme is also the optimal solution to $(P_1)$ (see Theorem 3.1 in \cite{BeckerS2010}). However, the number of iterations taken by the above scheme to convergence depends on the choice of $\mu$ and $\xx_0$. In practice, their values can be determined iteratively by the same continuation and ravine step techniques as previously described in FISTA.


\section{Augmented Lagrangian Methods}
\label{sec:ALM}
In the previous sections, we have seen the utility of Lagrange multipliers in $\ell_1$-min. In optimization, its basic idea is to eliminate \emph{equality constraints} by adding a suitable penalty term to the cost function that assigns a very high cost  to points outside the feasible set. In this section, we propose another special class of first-order methods called \emph{augmented Lagrangian methods} (ALM) to develop fast and scalable algorithms for both the standard $(P_1)$ and the CAB problem~\eqref{eq:CAB}. 

\subsection{Applying ALM to the primal problems}
Using the same notation from $(P_1)$, let $g(\xx) = \|\xx\|_1$ and $h(\xx) =\bb-A\xx$. Since both $g$ and $h$ are continuous, convex functions in $\xx$, we may assume $(P_1)$ has a unique global minimum. Hence, the following modified cost function, with an additional penalty term,
\begin{equation}
\min_\xx \, g(\xx) + \frac{\xi}{2}\|h(\xx)\|_2^2\quad \subj\quad h(\xx) = 0
\label{eq:augmented-penalty}
\end{equation}
has the same optimal solution as $(P_1)$, say $\xx^*$, for any $\xi > 0$. The quadratic penalty is preferred for its smoothness property, although other kinds of penalty functions are also plausible. Consider the Lagrangian of \eqref{eq:augmented-penalty} given by
\begin{equation}
\L_\xi(\xx,\bt) = g(\xx) + \frac{\xi}{2}\|h(\xx)\|_2^2 + \bt^T h(\xx),
\end{equation}
where $\bt \in \Re^m$ is a vector of Lagrange multipliers. $\lag_\xi(\cdot, \cdot)$ is called the {\it augmented Lagrangian function} of \eqref{eq:l1}. It has been shown in \cite{BertsekasD2003} that there exists $\bt^* \in \Re^m$ (not necessarily unique) and $\xi^* \in \Re$ such that
\begin{equation}
\xx^* = \arg\min_\xx \, \lag_\xi(\xx,\bt^*) \quad \forall \, \xi > \xi^*.
\label{eq:auglag}
\end{equation}
Thus, it is possible to find the optimal solution to $(P_1)$ by minimizing the augmented Lagrangian function $\L_\xi(\xx,\bt)$. Using the \emph{method of multipliers} \cite{BertsekasD2003}, a basic iterative scheme to solve \eqref{eq:auglag} is given by
\begin{equation}
\left \{
\begin{array}{lll}
\xx_{k+1} & = & \arg\min_{\xx} \, \lag_{\xi_k} (\xx,\bt_k)\\
\bt_{k+1} & = & \bt_k + \xi_k \, h(\xx_{k+1}) \\
\end{array}
\right . ,
\label{eq:alm}
\end{equation}
where $\left\{\xi_{k}\right\}$ is a predefined positive sequence. The fundamental convergence result of the above scheme states that $\{\xx_k\}$ and $\{\bt_k\}$ converge to $\xx^*$ and $\bt^*$, respectively, provided that $\{\bt_k\}$ is a bounded sequence and $\{\xi_k\}$ is sufficiently large after a certain index. Furthermore, the convergence rate is linear as long as $\xi_k > \xi^*$, and superlinear if $\xi_k \rightarrow \infty$ \cite{BertsekasD2003}.

Here, we point out that the choice of $\{\xi_k\}$ is problem-dependent. As shown in \cite{BertsekasD2003}, increasing $\xi_k$ increases the ill-conditionness or difficulty of minimizing $\lag_{\xi_k} (\xx,\bt_k)$, and the degree of difficulty depends on the condition number of $\nabla_{xx}^2 \lag_{\xi_k} (\xx_k,\bt_k)$. Thus, for experiments on synthetic data, we let $\xi_k \rightarrow \infty$ for better convergence rate; for experiments on real face data, we use a fixed $\xi_k \equiv \xi, \forall k$, to alleviate the difficulty.

Finally, it is easy to see that for $(P_1)$ the subproblem $\xx_{k+1} = \arg\min_{\xx} \, \lag_{\xi_k} (\xx,\bt_k)$ has the same form as $(QP_{\lambda})$, hence can be readily solved by many algorithms we have mentioned so far. In this paper we use FISTA for its simplicity and efficiency. The complete ALM algorithm for solving the primal $\ell_1$-min problem is referred to as \emph{Primal ALM} (PALM) in this paper. 


Next, we extend the ALM algorithm to solving the CAB problem \eqref{eq:CAB} for face recognition. We first write down the augmented Lagrangian function for this problem:
\begin{equation}
\L_\xi(\xx,\ee, \bt) = \|\xx\|_1 + \|\ee\|_1 + \frac{\xi}{2}\|\bb - A\xx - \ee\|_2^2 + \bt^T (\bb - A\xx - \ee),
\end{equation}
where we choose $\xi = 2m/\|\bb\|_1$ as suggested in  \cite{YangJ2009}. Here, the key for designing an efficient algorithm is to explore the special structure of the data matrix $B = [A, \ I]$ by computing $\xx$ and $\ee$ separately in each iteration:
\begin{equation}
\left \{
\begin{array}{lll}
\ee_{k+1} & = & \arg\min_{\ee} \, \lag_{\xi} (\xx_k,\ee, \bt_k)\\
\xx_{k+1} & = & \arg\min_{\xx} \, \lag_{\xi} (\xx,\ee_{k+1}, \bt_k)\\
\bt_{k+1} & = & \bt_k + \xi (\bb - A\xx_{k+1} - \ee_{k+1}) \\
\end{array}
\right .,
\label{eq:alm-CAB}
\end{equation}
where the subproblem for $\ee$ has a closed-form solution, and the subproblem for $\xx$ can be solved via FISTA. This algorithm is summarized in Algorithm \ref{alg:alm_rec}.
\footnote{Using the special structure of dictionary $[A, I]$, other $\ell_1$-min algorithms previously discussed can be similarly customized in face recognition.}
\begin{algorithm}[h]
\caption{Primal Augmented Lagrangian Method (PALM) for CAB} \label{alg:alm_rec}
\begin{algorithmic}[1]
\begin{small}
\STATE {\bf Input:} $\bb \in \Re^m$, $A \in \Re^{m \times n}$,
$\xx_1 = \mathbf{0}$, $\ee_1 = \bb$, $\bt_1 =
\mathbf{0}$.
\WHILE{not converged ($k = 1,2,\ldots$)}
\STATE $\ee_{k+1} \leftarrow \textup{shrink}(\bb - A\xx_k +
\frac{1}{\xi}\bt_k, \frac{1}{\xi})$;
\STATE $t_1\leftarrow 1$, $\zz_1 \leftarrow \xx_k$, $\ww_1 \leftarrow \xx_k$;
\WHILE{not converged ($l = 1,2,\ldots$)}
\STATE $\ww_{l+1} \leftarrow \textup{shrink}(\zz_l +
\frac{1}{L}A^T(\bb - A\zz_l - \ee_{k+1} +
\frac{1}{\xi}\bt_k), \frac{1}{\xi L})$;
\STATE $t_{l+1} \leftarrow \frac{1}{2}( 1 +
\sqrt{1+4t_l^2})$;
\STATE $\zz_{l+1} \leftarrow \ww_{l+1} + \frac{t_l - 1}{t_{l+1}}(\ww_{l+1} - \ww_l)$;
\ENDWHILE
\STATE $\xx_{k+1} \leftarrow \ww_{l}$,  \; $\bt_{k+1} \leftarrow \bt_k + \xi (\bb - A\xx_{k+1} - \ee_{k+1})$;
\ENDWHILE \STATE
{\bf Output:} $\xx^* \leftarrow \xx_k, \ee^* \leftarrow \ee_k$.
\end{small}
\end{algorithmic}
\end{algorithm}

In the literature, ALM algorithms have also been widely used in signal processing, and more recently in compressive sensing applications \cite{YangJ2009, GoldsteinO09, Afonso2010-TIP, Afonso2011-TIP}. Among all the existing works, the Alternating Direction Method (ADM) \cite{YangJ2009} essentially has the same form as our algorithm. The major difference is that \cite{YangJ2009} would approximate the solution to the subproblem for $\xx$ in \eqref{eq:alm-CAB} by computing only one iteration of the FISTA algorithm. Although this inexact ADM is guaranteed to converge, it only works well when the problem is well-conditioned. We have observed that it converges very slowly on real face data, hence is not suitable for our purpose.

\subsection{Applying ALM to the dual problems}

The principles of ALM can be also applied to the dual problem of $(P_1)$:
\begin{equation}
\max_{\yy}\, \bb^T\yy \quad \mbox{subj. to}\quad A^T\yy \in \mathbf{B}_1^{\infty},
\end{equation}
where $\mathbf{B}_1^{\infty} = \{\xx\in \Re^n:
\|\xx\|_{\infty}\leq 1\}$. The associated augmented Lagrangian function is given by
\begin{equation}
\begin{array}{c}
\min_{\yy, \zz}\, -\bb^T \yy - \xx^T(\zz-A^T\yy) + \frac{\beta}{2}\|\zz-A^T\yy\|_2^2 \\ \subj \quad \zz \in \mathbf{B}_1^{\infty}.
\end{array}
\end{equation}
Here, $\xx$ is the Lagrange multiplier for the dual problem. Since it is difficult to solve the above problem simultaenously w.r.t. $\yy$, $\xx$ and $\zz$, we again adopt an alternation strategy, where we iteratively minimize the cost function with respect to one of the variables while holding the rest constant.

On one hand, given $(\xx_k,\yy_k)$, the minimizer
$\zz_{k+1}$ with respect to $\zz$ is given by
\begin{equation}
\zz_{k+1} = \mathcal{P}_{\mathbf{B}_1^{\infty}}(A^T\yy_k + \xx_k / \beta),
\end{equation}
where $\mathcal{P}_{\mathbf{B}_1^{\infty}}$ represents
the projection operator onto $\mathbf{B}_1^{\infty}$. On the other hand,
given $(\xx_k,\zz_{k+1})$, the minimization
with respect to $\yy$ is a least squares problem, whose solution is given by the solution to the following equation:
\begin{equation}
\beta AA^T\yy = \beta A \zz_{k+1} - (A\xx_k - \bb).
\label{eqn:dual_alm}
\end{equation}
Suppose that $AA^T$ is
invertible, we can directly use its inverse to solve \eqref{eqn:dual_alm}. Finally, for the CAB problem \eqref{eq:CAB}, one can simply replace $A$ with $B = [A,\ I]$, and $\xx$ with $\ww = [\xx ; \ee]$, resulting in the DALM algorithm for face recognition as summarized in Algorithm~\ref{alg:alm_rec_dual}. Note that since all the subproblems are solved exactly, the convergence of the dual algorithm is guaranteed. 


Meanwhile, it is pointed out in \cite{YangJ2009} that the matrix inversion step can be computationally expensive. Therefore, one can approximate the solution with one step of the conjugate gradient algorithm in the $\yy$ direction at each iteration. However, we find that while this heuristic works fine on synthetic data when $A$ is a random Gaussian matrix, it does not work for the dictionary $A$ formed by real face images.

Finally, we provide some general comments about the difference between the primal and dual ALM algorithms and its implications in face recognition. While both algorithms are guaranteed to solve the $\ell_1$-min problem in theory, their efficiency can be very different in different real-world applications. As we mentioned before, for face recognition problem and particularly solving Eq.~\eqref{eq:CAB}, it is crucial for an algorithm to scale well with thousands or even more subjects (i.e., $n$), while the dimension of each face image (i,e., $m$) remains relatively a constant. For PALM, the computational time is dominated by matrix-vector multiplication in Step 6 of Algorithm~\ref{alg:alm_rec}, whose complexity is $O(n^2)$, whereas the most computational step in DALM is Step 4 of Algorithm~\ref{alg:alm_rec_dual} with $O(m^2+mn)$ complexity. This is also evidenced in our experiment results in Section~\ref{sec:experiment}. Therefore, DALM should be preferred in the case. On the contrary, for face alignment problem \eqref{eqn-align-simple}, since the dictionary only has a small number of columns, we will see later that PALM is much faster than DALM in practice.

\begin{algorithm}[t!]
\caption{Dual Augmented Lagrangian Method (DALM) for CAB} \label{alg:alm_rec_dual}
\begin{algorithmic}[1]
\begin{small}
\STATE {\bf Input:} $\bb \in \Re^m$, $B = [A,\ I] \in \Re^{m \times (n+m)}$,
$\ww_1 = \mathbf{0}$, $\yy_1 =
\mathbf{0}$.
\WHILE{not converged ($k = 1,2,\ldots$)}
\STATE $\zz_{k+1} = \mathcal{P}_{\mathbf{B}_1^{\infty}}(B^T\yy_k + \ww_k / \beta)$;
\STATE $\yy_{k+1} = (BB^T)^{-1}(A \zz_{k+1} - (B\ww_k - \bb) /\beta)$;
\STATE $\ww_{k+1} = \ww_{k} - \beta(\zz_{k+1} - A^T\yy_{k+1})$;
\ENDWHILE \STATE
{\bf Output:} $\xx^* \leftarrow \ww_k[1:n], \ee^* \leftarrow \ww_k[n+1:n+m], \yy^* \leftarrow \yy_k$.
\end{small}
\end{algorithmic}
\end{algorithm}




\section{Experiments}
\label{sec:experiment}
In this section, we validate and benchmark the performance of the primal and dual ALM algorithms against an extensive list of seven state-of-the-art $\ell_1$-min solvers. The other algorithms involved in the comparison are PDIPA, TNIPM/L1LS, Homotopy, FISTA, TFOCS, SESOP-PCD, and AMP. All experiments are performed in MATLAB on a Mac Pro with two 2.66\,GHz 6-Core Intel Xeon processors and 24\,GB of memory.

The experiment consists of three sets of benchmark settings. The first benchmark compares the accuracy and speed of the
algorithms for solving the generic $\ell_1$-min problems. Here the sparse source signal
and the underdertermined linear system are randomly generated based on Gaussian distribution. The second benchmark measures the face recognition accuracy via the CAB model in a real-world face recognition scenario, where the training and query images are taken from a public face recognition database. The last benchmark compares the performance of an image alignment problem in face recognition, where the query images may contain face pose variations or image registration error in 2-D.

\noindent{\bf Metrics of performance.} A primary measure of performance in this paper is the relative error $r_k(\xx)$ as a function of CPU time $t_k$ after $k$ iterations:
$$
r_k(\xx) = \frac{\|\xx_k - \xx_0\|_2}{\|\xx_0\|_2},
$$
where $\xx_k$ is the estimate after
$k$ iterations. For experiments on the real face data where the ground truth $\xx_0$ is not known in advance, we use Homotopy to determine it before running the comparison experiments, as Homotopy is able to solve $(P_1)$ exactly (to an error level comparable to the machine precision). Note that the same strategy was also used in \cite{LorisI2009}.

\noindent{\bf Unifying the optimization problems}. A key to a fair comparison is to ensure that all the
algorithms are solving the same optimization problem. That is,
one should not confuse the problem about how well an
$\ell_1$-min model fits the compressive sensing applications (for example,
achieving high face recognition rates) with the problem about
how well an $\ell_1$-min algorithm does in finding the solution
of the optimization problem assigned to it. In this paper, we are only interested in the latter one and have restricted our attention to the basic $(P_1)$ problem. Among all the algorithms, PDIPA, Homotopy, TFOCS, AMP and ALM are designed to solve $(P_1)$, while L1LS, FISTA and SESOP-PCD solve the unconstrained basis pursuit problem $(QP_{\lambda})$: 
\begin{displaymath}
\quad \xx_{\lambda}^* = \arg \min_\xx \, \frac{1}{2}\|\bb - A\xx\|_2^2 + \lambda \|\xx\|_1.
\end{displaymath}

\begin{figure}[t]
\centering
\includegraphics[height=1.25in]{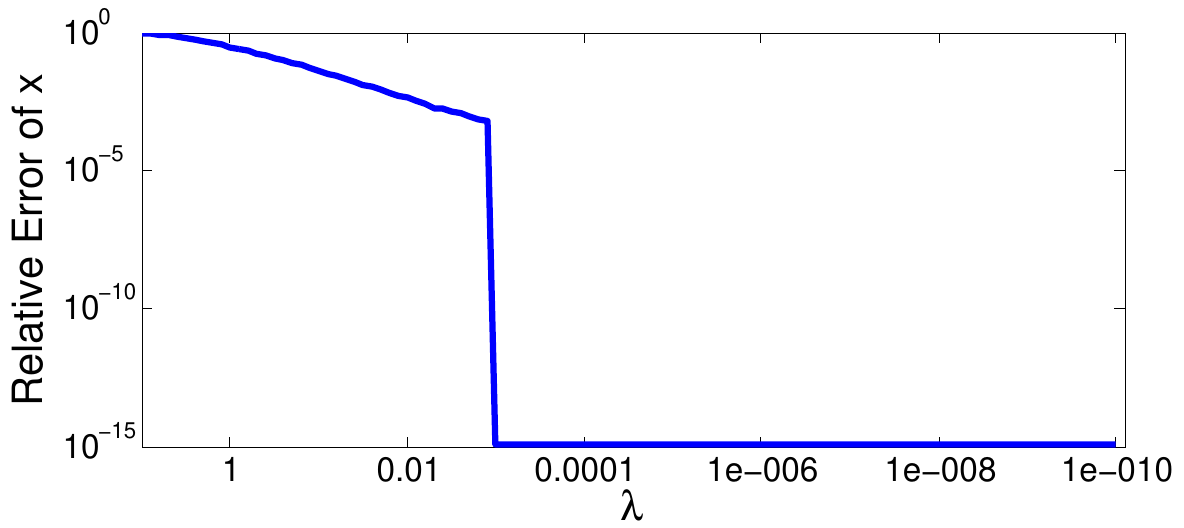}
\caption{Relative error of $\xx$ as a function of $\lambda$ for Homotopy algorithm. } \label{fig:Homotopy}
\end{figure}

Although it is obvious that as $\lambda \rightarrow 0$, the solution of $(QP_{\lambda})$ converges to that of  $(P_1)$, it is not practical to directly set $\lambda=0$ in numerical computations. Fortunately, we recall in the Homotopy algorithm that the solution path $\{\xx_{\lambda}^*, \lambda \in [0,\infty)\}$ is a piecewise linear path, with a finite number of vertices. This suggests that for any single instance of the $\ell_1$-min problem, there always exists a $\bar{\lambda} > 0$ such that for any $\lambda < \bar{\lambda}$ the solution of $(QP_{\lambda})$ is also the solution of $(P_1)$. This is illustrated in Figure~\ref{fig:Homotopy} by a sharp drop in the relative error of $\xx$ (to a level comparable to the machine precision) at certain positive value of $\lambda$. In addition, note that the value of $\bar{\lambda}$ can be obtained by the Homotopy algorithm without any extra cost, as it finds the solution $x_{\lambda}^*$ for all $\lambda\in [0, \infty)$. Therefore, we can safely choose any $\lambda^* < \bar{\lambda}$ for L1LS, FISTA and SESOP-PCD in our experiments to ensure fair comparison, and we find that fixing $\lambda^* = 10^{-6}$ suffices in all the experiments.


\noindent{\bf Warm-start strategy.} A commonly used acceleration technique for algorithms that solve $(QP_{\lambda})$ is the so-called \emph{warm-start} strategy. The idea is that, in order to find the solution of $(QP_{\lambda})$ for $\lambda =\lambda^*$, one solves a series of $(QP_{\lambda})$ problems with parameters $\lambda_0 > \lambda_1 > \cdots > \lambda_N = \lambda^*$, and in each step the previous solution $\xx_{\lambda_{j-1}}^*$ is used to initialize $\xx_{\lambda_j}^*$. 

To further simplify the warm-start procedure, a \emph{fixed-point continuation method} has been considered in \cite{FigueiredoM2007, HaleE2007}. In this method, instead of solving $(QP_{\lambda})$ for each $\lambda_j$ exactly, one starts with $\lambda = \lambda_0$ and decreases it geometrically after each iteration $\lambda_{k+1} = \rho \lambda_k$ until it reaches $\lambda^*$. In our experiment, we found that solving a series of $(QP_{\lambda})$ problems works better for L1LS and SESOP-PCD, where as the fixed-point continuation method is more effective for FISTA. Therefore, we choose a series of  $(QP_{\lambda})$ problems with $\lambda = \{10^{-1}, 10^{-2}, \ldots, 10^{-6}\}$ for L1LS and SESOP-PCD, and use the fixed-point continuation method with $\rho = 0.95$ for FISTA.

\subsection{Synthetic data, the noise-free case}

\begin{figure*}[t]
\centering
\begin{tabular}{ccc}
\hspace{-2mm}\includegraphics[height=1.4in]{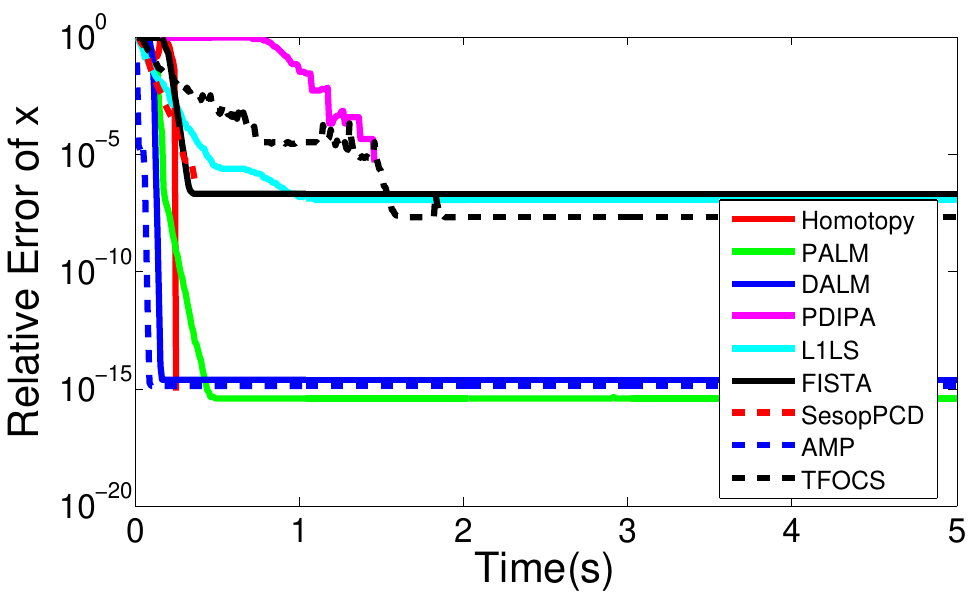}&
\hspace{-2mm}\includegraphics[height=1.4in]{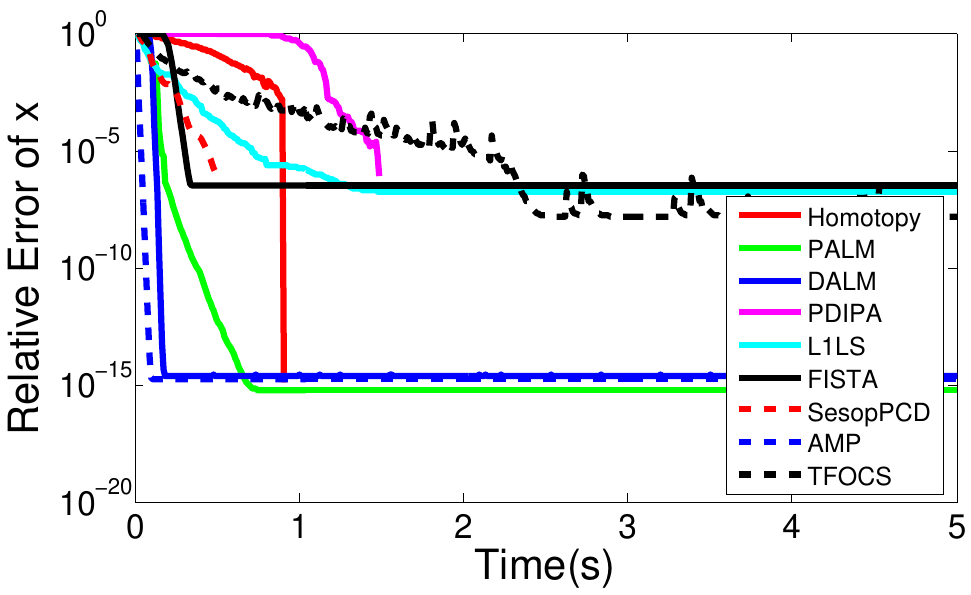} &
\hspace{-2mm}\includegraphics[height=1.4in]{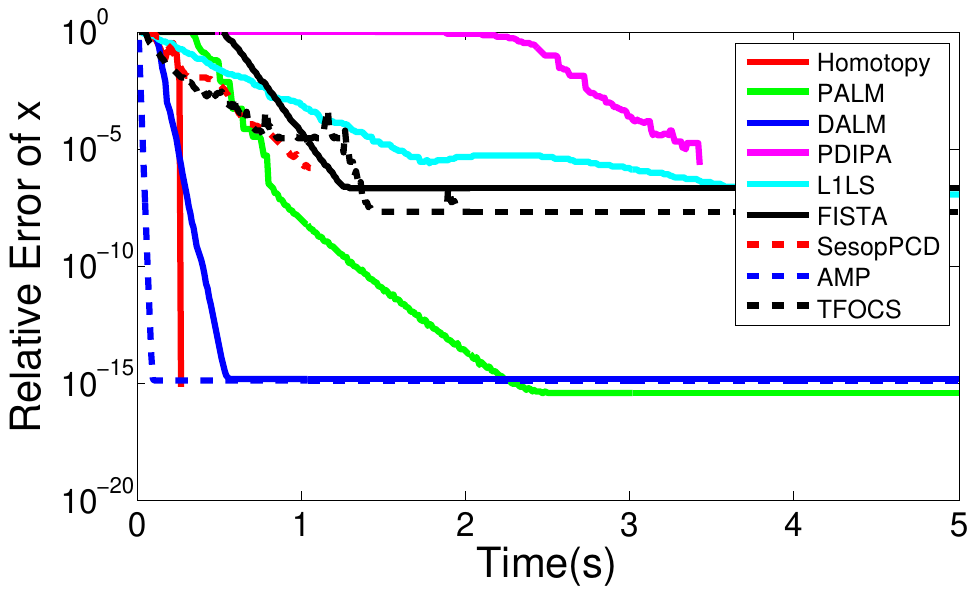}\\
(a) $m=800, n=1000, d = 100$ & (b) $m=800, n=1000,  d = 200$ & (c) $m=800, n=2000,  d = 100$\\
\end{tabular}
\caption{{\bf Synthetic data experiment:} Relative error of $\xx$ as a function of CPU time.} \label{fig:synthetic}
\end{figure*}

In the first experiment, we compare the time taken to solve $(P_1)$
by the nine algorithms described earlier. We generate the
observation matrix $A$ of size $m \times n$ ($m<n$), such that
each entry in the matrix is independent and identically distributed (i.i.d.) 
Gaussian. In addition, we normalize each column to
have unit $\ell_2$-norm. The observation $\bb$ is computed by $A\xx_0$,
where $\xx_0$ is a sparse vector with $\|\xx_0\|_0 = d$. The support of $\xx_0$ is also chosen
at random, and the nonzero entries of $\xx_0$ are i.i.d.
according to a uniform distribution in the interval $[-10,10]$. We fix $m = 800$, and for different choices of $n$ and $d$, we compute the relative errors in estimating $\xx_0$ as a function of CPU time using all the algorithms in question. Figure~\ref{fig:synthetic}(a)-(c) shows the averaged relative errors over 20 trials. 

We observe in all three plots that AMP is the fastest algorithm in all the cases, followed by DALM. Moreover, AMP, PALM, DALM and Homotopy are the only methods that achieve near-machine precision ($r_k(\xx) \leq 10^{-10}$) in solving $(P_1)$. This differentiates them from the other methods.

\begin{figure}[t]
\centering
\begin{tabular}{cc}
\hspace{-4mm}\includegraphics[height=1.35in]{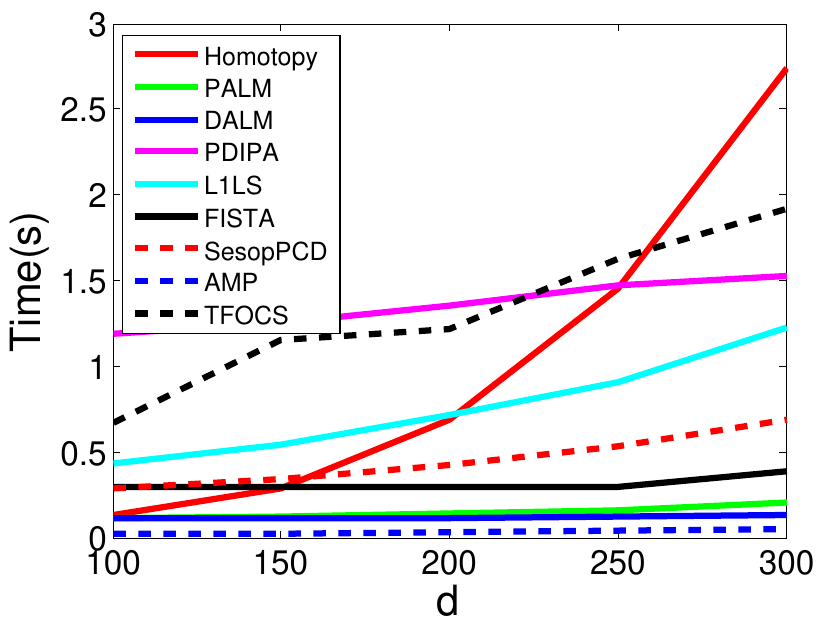} &
\hspace{-4mm}\includegraphics[height=1.35in]{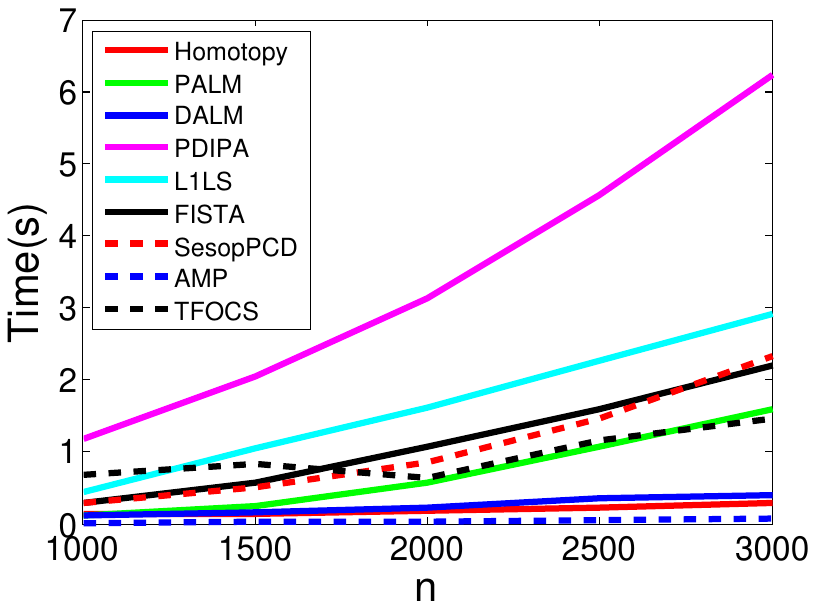}\\
\end{tabular}
\caption{{\bf Synthetic data experiment:} Time to reach $10^{-5}$ accuracy as a function of sparsity level $d$  {\bf (left)} and dimension $n$  {\bf (right)}.} \label{fig:synthetic-time}
\end{figure}

Next, Figures~\ref{fig:synthetic-time}(left) shows the time taken by each algorithm to achieve a prescribed tolerance $r_k(\xx) = 10^{-5}$ with various sparsity levels $d$. As one can see, all algorithms slow down when $d$ in the true solution $\xx$ increases. However, it affects most prominently Homotopy, as the time taken increases by more than a factor of 10.

Finally, in Figure~\ref{fig:synthetic-time}(right), the computational time of PDIPA increases dramatically as the problem dimension $n$ increases. However, as expected, Homotopy is not sensitive to the increase of $n$. More interestingly, compared to the primal algorithms (PDIPA, L1LS, FISTA, SesopPCD, AMP and PALM), the dual algorithms (DALM and TFOCS) are much less sensitive to $n$. In fact, dual algorithms and Homotopy are the only ones which scale linearly in $n$. 

\subsection{Real face data, the recognition experiment}

In this case, the observation satisfies the CAB model: $\bb =
A\xx_0 + \ee_0$, where both $\xx_0$ and $\ee_0$ are sparse
vectors. We first note that AMP is designed specifically for $\ell_1$-min problems when $A$ is a random Gaussian matrix, hence is excluded in this experiment. In fact, we have seen that the algorithm simply does not converge with an estimation error going to infinity when applied to real face data. Secondly, all the algorithms in this section have been carefully modified to take into account the special data structure of the CAB model: $B = [A, I]$, and the sparse vectors $\xx$ and $\ee$ are treated separately in their respective routines.

The performance of the $\ell_1$-min algorithms is benchmarked on the CMU Multi-PIE database \cite{GrossR2006}. A subset of 249 subjects from the database (Session 1) is used for this experiment. Each subject is captured under 20 different illuminations with a frontal pose. The images are then manually aligned and cropped, and down-sampled to $40\times 30$ pixels. Out of the 20 illuminations for each subject, we choose $l$ illuminations as the training, resulting a measurement matrix $A$ of size $1200\times 249l$. Further, we randomly choose 100 images from the remaining images as test images. Finally, a certain fraction $p \in [0,1)$ of image pixels are randomly corrupted by a uniform distribution between $[0,255]$.

\begin{figure}[t]
\centering
\begin{tabular}{cc}
\hspace{-4mm}\includegraphics[height=1.5in]{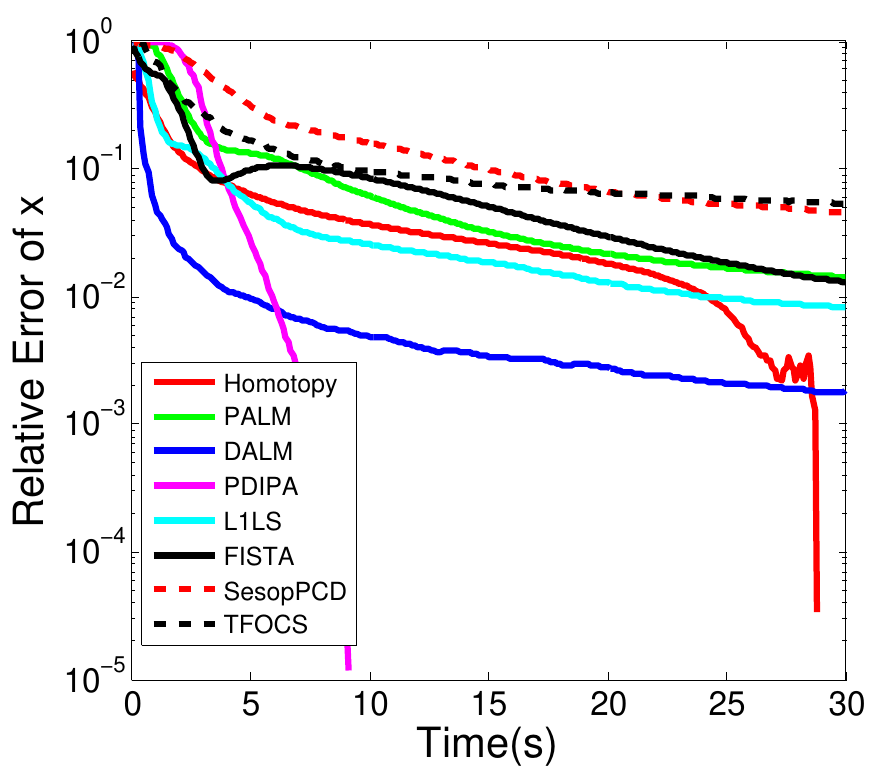}&
\hspace{-4mm}\includegraphics[height=1.5in]{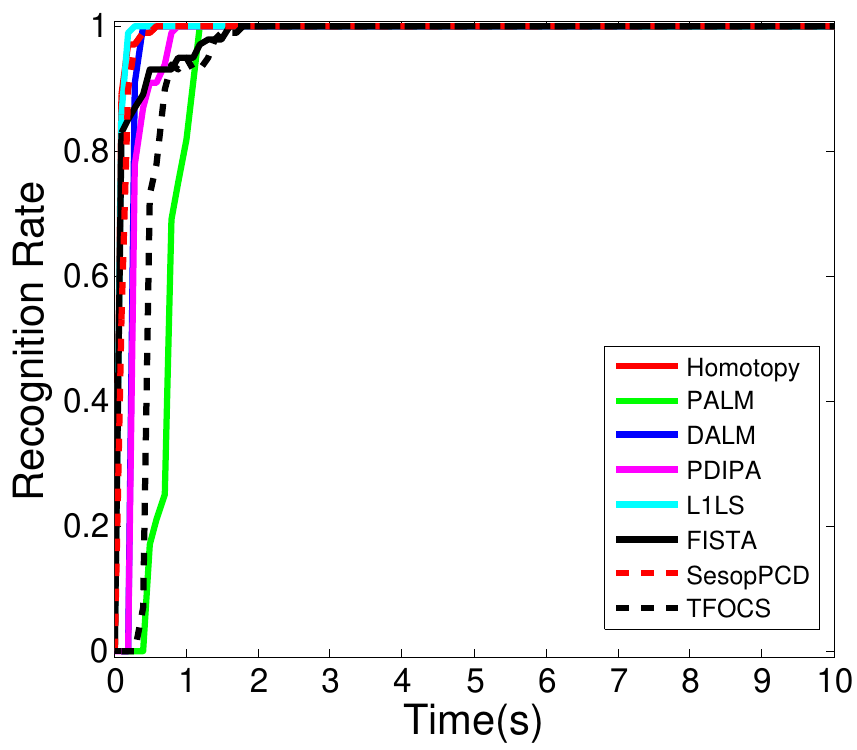}\\
\hspace{-4mm}\includegraphics[height=1.5in]{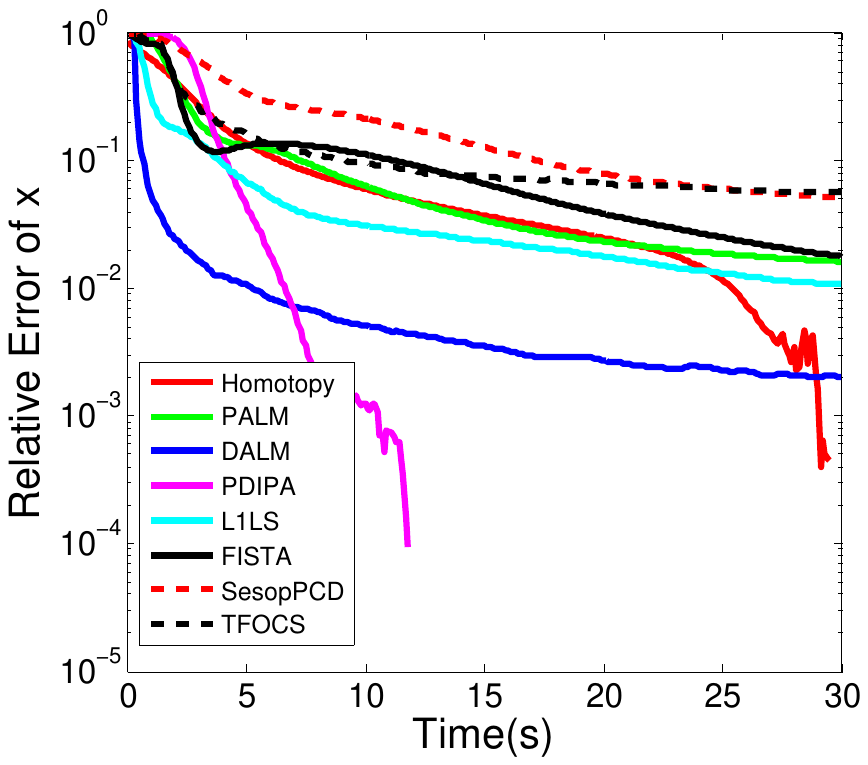} &
\hspace{-4mm}\includegraphics[height=1.5in]{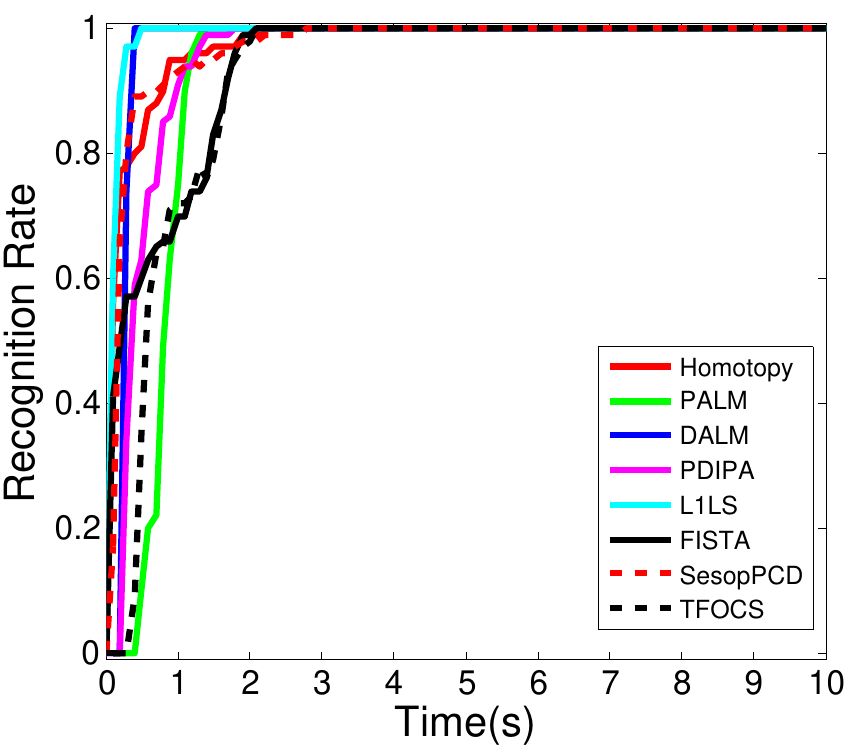} \\
\hspace{-4mm}\includegraphics[height=1.5in]{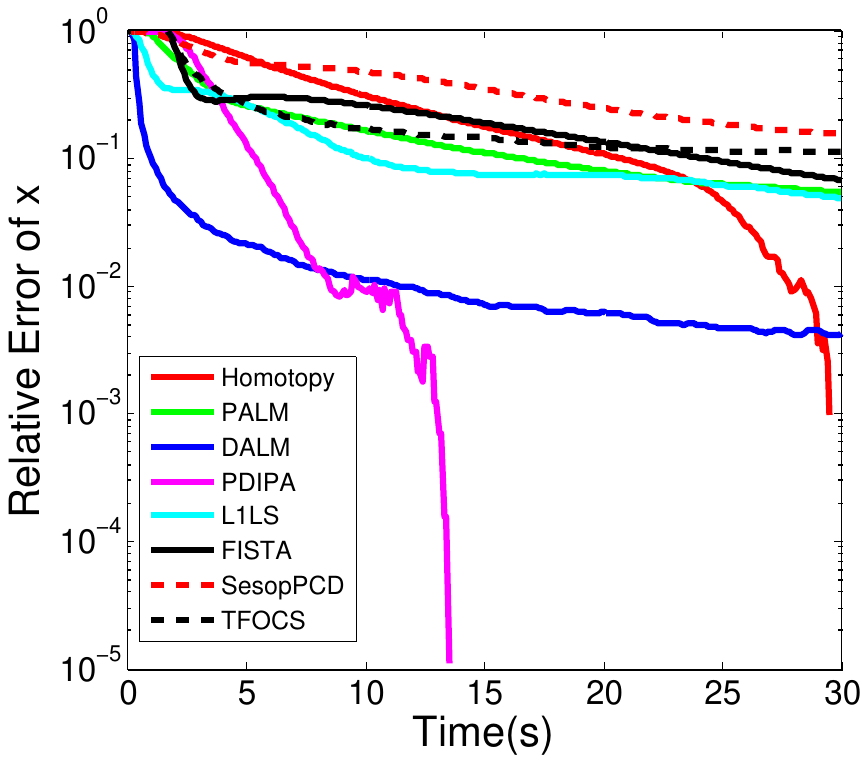}&
\hspace{-4mm}\includegraphics[height=1.5in]{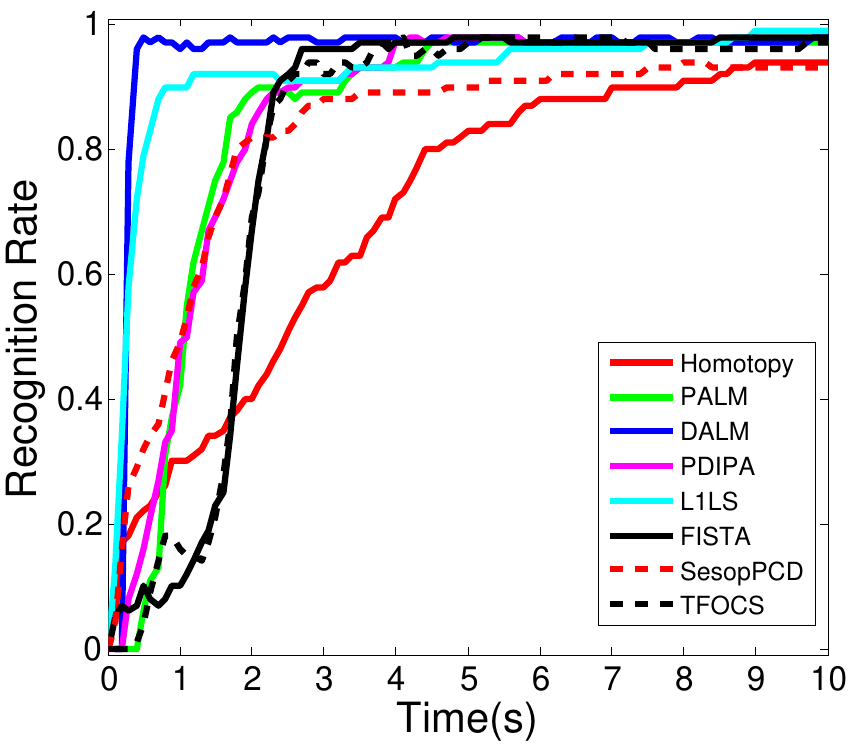}\\
(a) Relative error of $\xx$ & (b) Recognition rate
\end{tabular}
\caption{Comparison of $\ell_1$-min algorithms on Multi-PIE with different fractions of corrupted entries. {\bf From top to bottom:} $p = 0.2, 0.4, 0.6$.} \label{fig:recognition-p}
\end{figure}

To measure the performance of each algorithm, we again compute the relative error in estimating $\xx$ after each iteration. The ground truth $\xx_0$ is obtained by Homotopy. We also compute the recognition rate after each iteration, where we use the Sparsity Concentration Index (SCI) proposed in \cite{WrightJ2009-PAMI} as the classification criterion. That is, we assign a test image to subject $i^*$ if 
\begin{equation}
i^* = \arg \max_i \frac{C \cdot \|\delta_i(\xx_k)\|_1/\|\xx_k\|_1 - 1}{C-1},
\end{equation}
where $C$ is the number of subjects, and $\delta_i(\xx_k)$ is a projection that only keeps the entries associated with subject $i$.

\begin{figure}[t]
\centering
\begin{tabular}{cc}
\hspace{-4mm}\includegraphics[height=1.5in]{speed_p40_n249-r.pdf} &
\hspace{-4mm}\includegraphics[height=1.5in]{rec_p40_n249-r.pdf} \\
\hspace{-4mm}\includegraphics[height=1.5in]{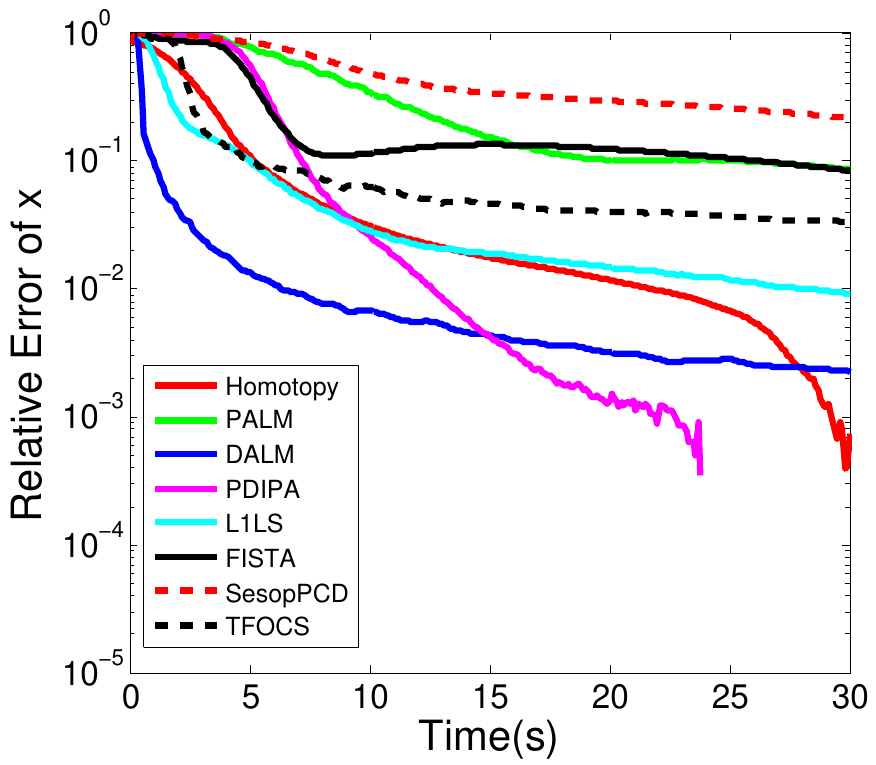} &
\hspace{-4mm}\includegraphics[height=1.5in]{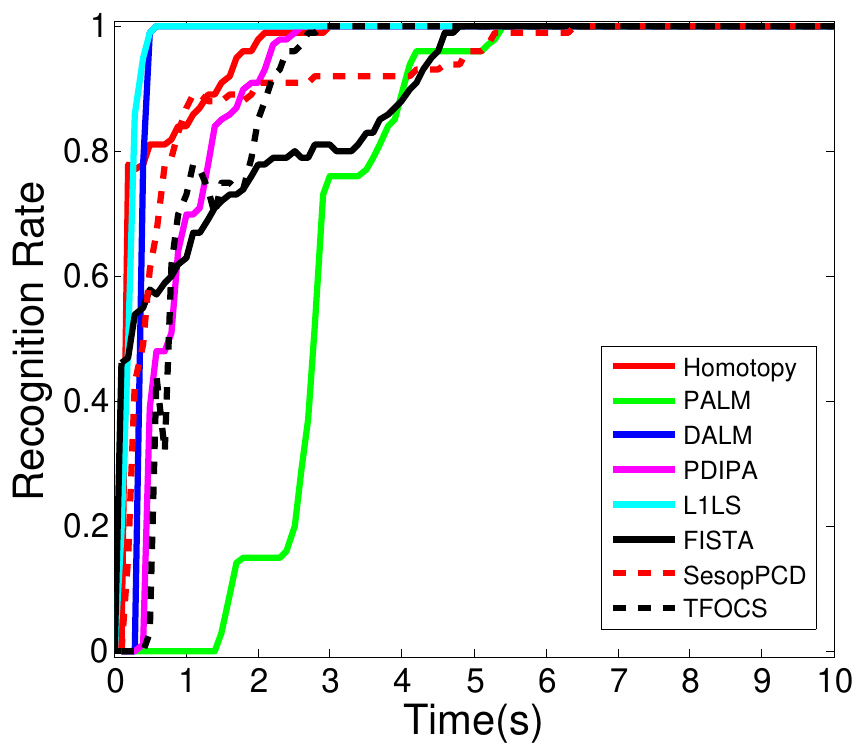} \\
\hspace{-4mm}\includegraphics[height=1.5in]{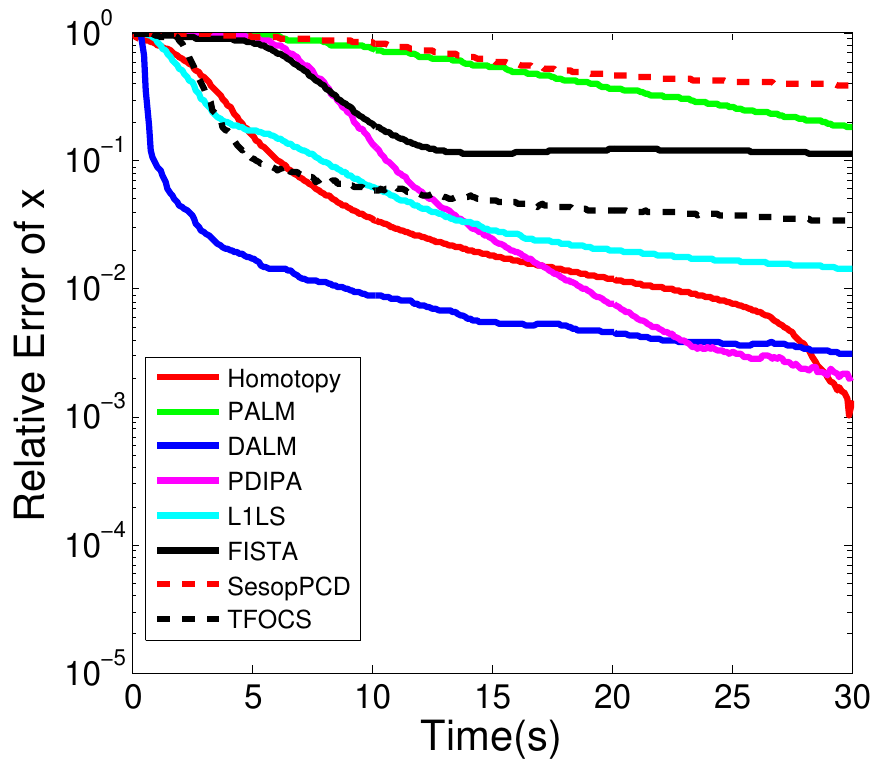} &
\hspace{-4mm}\includegraphics[height=1.5in]{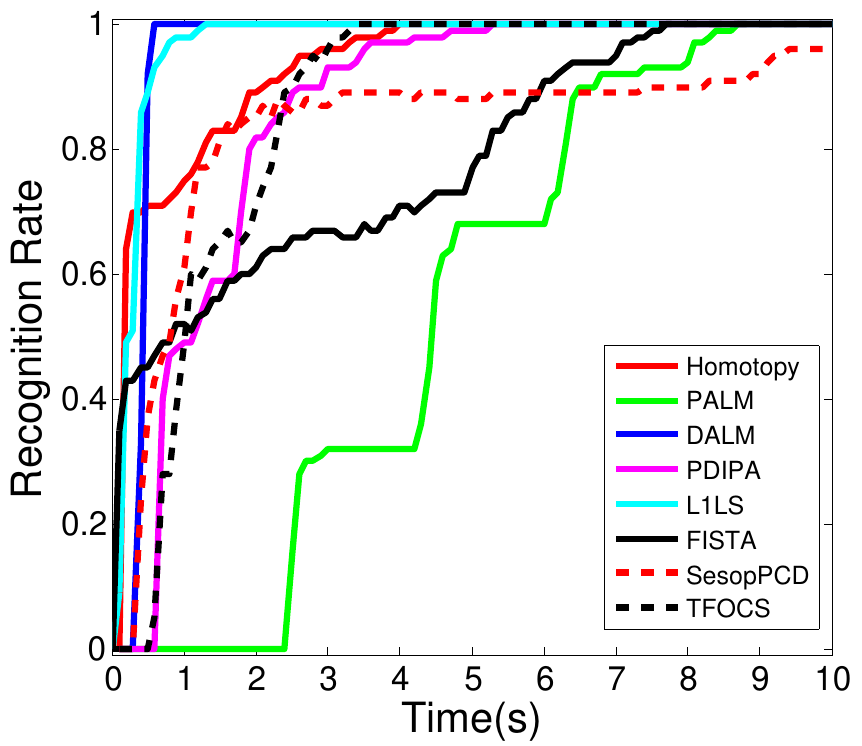} \\
(a) Relative error of $\xx$ & (b) Recognition rate
\end{tabular}
\caption{Comparison of $\ell_1$-min algorithms on Multi-PIE with different numbers of images per subject. {\bf From top to bottom:} $l= 7, 11, 15$.} \label{fig:recognition-k}
\end{figure}

In Figure~\ref{fig:recognition-p}, we fix $l=7$ and show the performance of $\ell_1$-min algorithms with various values of $p$. As one can see in Figure~\ref{fig:recognition-p}(a), PDIPA and Homotopy achieve the highest accuracy in estimating $\xx_0$ for all cases within a fixed time limit (30 seconds), but the relative error for DALM decreases much faster than the other algorithms in the first several seconds. Because of this, as one can see in Figure~\ref{fig:recognition-p}(b), DALM achieves the best possible recognition rates in a short time. For instance, when $p=0.6$, DALM achieves 98\% recognition rate in 0.5 second, while it takes PDIPA 4.2 seconds to reach the same rate. In addition, similar to the synthetic data case, all the algorithms slow down as the number of nonzero entries of $\ee_0$ increases. However, this effect is very small for DALM, especially in terms of achieving the highest recognition rates. 

In Figure~\ref{fig:recognition-k}, we fix $p=0.4$ and show the performance of $\ell_1$-min algorithms with various values of $l$. In all cases,  DALM is the fastest one to achieve 100\% recognition rate. Also, as we discussed before, since DALM solves the dual problem of $(P_1)$, its efficiency is much less affected by the change of the size of the primal variable $\xx$, which makes it most suitable for face recognition applications with a large number of subjects. Another algorithm that performs well in this experiment is L1LS, but it is less accurate than DALM in estimating $\xx$, as shown in Figure~\ref{fig:recognition-k}(a).

In summary, DALM is the best among all tested for robust face recognition, as it achieves high recognition rates in our experiments and scales well for large-scale applications. 

\subsection{Real face data, the alignment experiment}
In this experiment, we modify each algorithm to solve the associated $\ell_1$-min problem of the image alignment problem whereby $\|\ww\|_1$ is no longer penalized as in \eqref{eqn-align-simple}. However, since Homotopy is designed specifically to solve the original $\ell_1$-min problems that would include $\|\ww\|_1$ in the objective function, it is further excluded from the experiment.

\begin{figure}[t]
\centering
\begin{tabular}{cc}
\includegraphics[height=1in,clip=true]{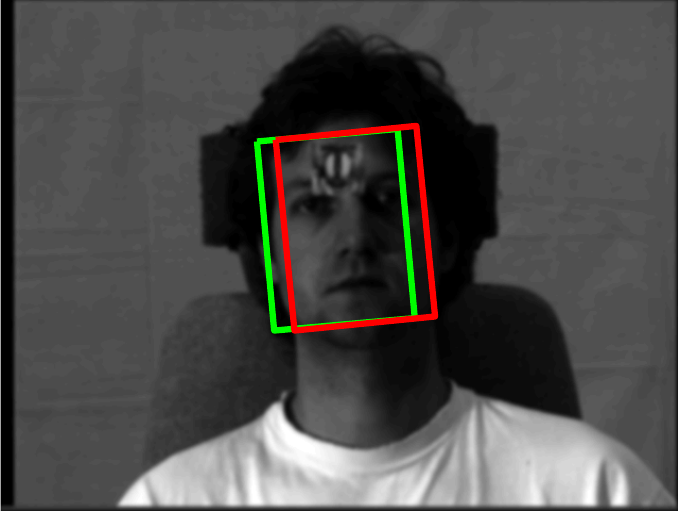} &
\hspace{-2mm}\includegraphics[height=1in,clip=true]{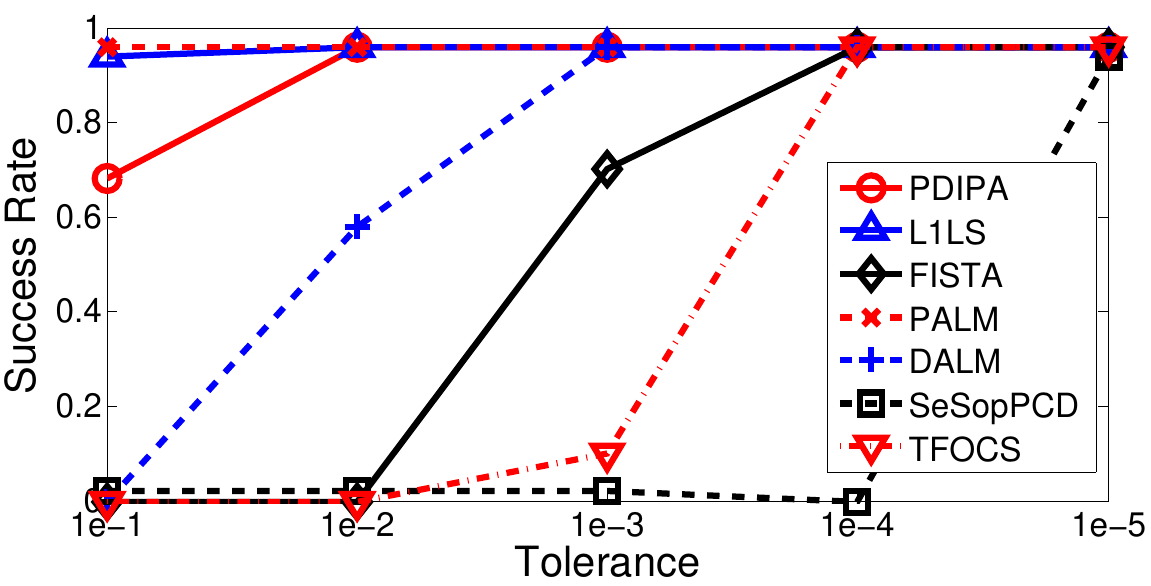} \\
\includegraphics[height =1in,clip=true]{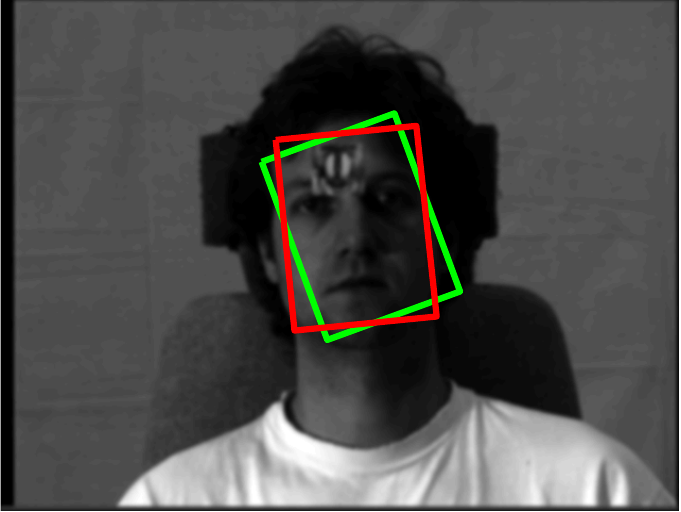} &
\hspace{-2mm}\includegraphics[height =1in,clip=true]{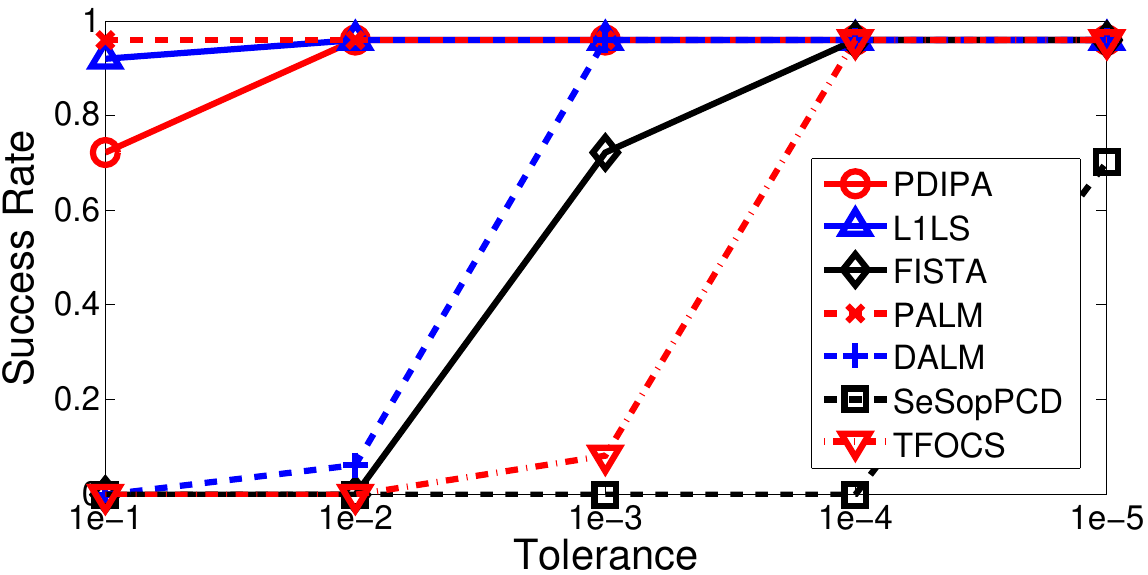} \\
(a) & (b)
\end{tabular}
\caption{Face alignment experiments. {\bf (a)} Two cases of
perturbation. The green boxes indicate the initial face location; the red boxes show the final position after face alignment. {\bf (b)} Rate of success as a function of the tolerance level.} \label{fig:alignment-expt}
\end{figure}

We again use the CMU Multi-PIE database to benchmark the $\ell_1$-min algorithms. In this experiment, the first 50 subjects from Session 1 are used. Our of 20 illuminations, seven are chosen as the training images\footnote{The training are illuminations $\{0,1,7,13,14,16,18\}$ of \cite{GrossR2006}.} and the illumination 10 is used as the query image. We down-sample the face region in the images to $40\times 30$ pixels. Moreover, to test the robustness of $\ell_1$-min algorithms to occlusion, we replace a randomly located block of size equal to $10\%$ of the face image with the image of a baboon (see Figure~\ref{fig:alignment-expt}). 

Here, we note that for the alignment experiments, the training set contains $n_i = 7$ images per subject, and we choose the transformation group $T$ to be the set of similarity transformations (therefore $q_i = 4$). So the number of columns in $B_i$ is just $n_i+q_i = 11$, while the number of rows $m = 40\times 30 = 1200$ in our experiments. A direct consequence of such highly overdetermined matrix $B$ is that algorithms operate in the primal space such as PDIPA and PALM are much more efficient than algorithms operate in the dual space such as DALM and TFOCS.

We consider two different types of misalignment, namely, translation and rotation. For translation, each test image is manually perturbed by 4 pixels along the $x$-axis in the canonical frame (with size $40\times 30$); for rotation, each test image is manually perturbed by $15$ in-plane degrees, as shown in Figure~\ref{fig:alignment-expt}(a). We stop the alignment process when (1) the difference between the final alignment is within 2 pixels of the ground truth in the original image frame ($640\times 480$ image size) or (2) a pre-defined maximum number of iterations is reached, and only consider an alignment successful if condition (1) is satisfied.

In addition, since now we have to solve a series of $\ell_1$-min problems \eqref{eqn-align-simple} for one alignment task, we need to specify the stopping criterion for each $\ell_1$-min instance. While different algorithms often adopt different stopping criteria in practice, for fair comparison, in this paper we compute the relative change of the estimate for each algorithm, $\|\ww_{k+1} - \ww_{k}\|_2/\|\ww_k\|_2$, and terminate the algorithm when the relative change is smaller than some prescribed value $tol$. 

It is easy to see that the smaller $tol$ is, the more accurate the $\ell_1$-min problem is solved at each iteration, and the more time each algorithm will take. So it is necessary to decide a good $tol$ for the alignment problem in practice. For this reason, we first show the success rate of each algorithm as a function of $tol$ in Figure~\ref{fig:alignment-expt}(b). As one can see, for each algorithm there exists certain $tol^*$ such that only when $tol<tol^*$ the algorithm achieves the best success rate. More surprisingly, the ranges of $tol$ for different algorithms to achieve the best success rate vary significantly in practice. For example, while PALM and L1LS achieve the best success rate even when $tol=10^{-1}$, FISTA and TFOCS only work well when $tol \leq10^{-4}$, which greatly limits their efficiency. Meanwhile, SESOP-PCD performs the worst in the experiment, failing to achieve the same success rate as the other algorithms.

\begin{table}[ht!]
\caption{Average time and iterations of $\ell_1$-min algorithms for face alignment. The fastest time is shown in bold.}
\label{tab:alignment}
\centering \small
\begin{tabular}{|c|c|c|c|c|c|}
\hline
\multirow{2}{*}{$tol$} & \multirow{2}{*}{Method} &  \multicolumn{2}{|c|}{Translation} &  \multicolumn{2}{|c|}{Rotation}\\
\cline{3-6}
& & Time (s) & Iterations & Time (s) & Iterations\\ 
\hline
\multirow{3}{*}{$10^{-2}$} & PALM  & {\bf 0.18} & 11.31 & {\bf 0.13} & 8.44 \\ & PDIPA  & 0.40 & 11.77 &  0.28 & 9.21 \\ & L1LS  & 2.71 & 11.60 &  1.96 & 8.75\\
\hline
\multirow{4}{*}{$10^{-3}$} & PALM  & {\bf 0.27} & 11.52 & {\bf 0.17} & 8.69 \\ & PDIPA  & 0.49 & 11.52 & 0.36 & 8.79 \\ & L1LS  & 3.76 & 11.50 & 2.84 & 8.71\\ & DALM & 5.44 & 14.35 & 6.29 & 14.79\\
\hline
\multirow{6}{*}{$10^{-4}$} & PALM  & 0.62  & 11.56 & {\bf 0.36} & 8.75 \\ & PDIPA  & {\bf 0.53} & 11.52 &  0.39 & 8.77 \\ & L1LS & 4.90 & 11.50 & 3.59 & 8.67\\ & DALM & 11.39 & 11.77 & 9.61 & 9.10\\ & FISTA & 8.75 & 11.60 &  6.62 & 8.85\\ & TFOCS & 249.92 & 14.46 & 232.78 & 12.85\\
\hline
\end{tabular}
\end{table}

Finally, we report the speed of $\ell_1$-min algorithms with various values of $tol$ in Table~\ref{tab:alignment}. Note that we only report the result of an algorithm with certain $tol$ if it achieves the best success rate in that case. As one can see, PALM, PDIPA and L1LS outperform the other algorithms in both translation and rotation experiments. Furthermore, for these three algorithms, the average number of iterations roughly remains constant for different $tol$. However, the computational time that each iteration costs increases as $tol$ decreases, so does the total time for the entire alignment task. This justifies the need of choosing the right $tol$ for each algorithm. Finally, for the same $tol$, PALM is the fastest among all algorithms, except for the case of translation with $tol=10^{-4}$. Therefore, we conclude that PALM is the best choice for this problem. 

\section{Conclusion}
\label{sec:conclusion}

In this paper, we have provided a comprehensive comparison of several popular $\ell_1$-min algorithms in a new practical scenario that has drawn a lot of attention in the signal processing and computer vision communities, namely, the sparse representation based robust face recognition applications. We have shown that the ALM algorithms compare favorably to other classical and accelerated sparse optimization methods, especially when applied to real face images. In particular, the dual ALM algorithm performs the best in the face recognition experiment, and scales well in terms of the number of subjects. Hence it is suitable for large-scale classification problems. Meanwhile, the primal ALM algorithm is the fastest method in solving the face alignment problem. Finally, we note that the performance of different numerical algorithms also depends on the programming language and the computer platform. For example, we have recently studied parallel implementation of the $\ell_1$-min problems on multi-core CPUs/GPUs. The interested reader is referred to \cite{ShiaV2011-Asilomar} for more details.


\section*{Acknowledgment}

The authors would like to express sincere appreciation to Dr. Zhouchen Lin at Peking University, Dr. Marc Teboulle at Tel-Aviv University, Dr.  Michael Zibulevsky at Technion - Israel Institute of Technology, and Dr. John Wright at Columbia University for their valuable comments.
\ifCLASSOPTIONcaptionsoff
  \newpage
\fi



\begin{thebibliography}{10}
\providecommand{\url}[1]{#1}
\csname url@samestyle\endcsname
\providecommand{\newblock}{\relax}
\providecommand{\bibinfo}[2]{#2}
\providecommand{\BIBentrySTDinterwordspacing}{\spaceskip=0pt\relax}
\providecommand{\BIBentryALTinterwordstretchfactor}{4}
\providecommand{\BIBentryALTinterwordspacing}{\spaceskip=\fontdimen2\font plus
\BIBentryALTinterwordstretchfactor\fontdimen3\font minus
  \fontdimen4\font\relax}
\providecommand{\BIBforeignlanguage}[2]{{%
\expandafter\ifx\csname l@#1\endcsname\relax
\typeout{** WARNING: IEEEtran.bst: No hyphenation pattern has been}%
\typeout{** loaded for the language `#1'. Using the pattern for}%
\typeout{** the default language instead.}%
\else
\language=\csname l@#1\endcsname
\fi
#2}}
\providecommand{\BIBdecl}{\relax}
\BIBdecl

\bibitem{YangA2010-ICIP}
A.~Yang, A.~Ganesh, S.~Sastry, and Y.~Ma, ``Fast $\ell_1$-minimization
  algorithms and an application in robust face recognition: a review,'' in
  \emph{Proceedings of the International Conference on Image Processing}, 2010.

\bibitem{CandesE2006-ICM}
E.~Cand{\`{e}s}, ``Compressive sampling,'' in \emph{Proceedings of the
  International Congress of Mathematicians}, 2006.

\bibitem{DonohoD2004}
D.~Donoho, ``For most large underdetermined systems of linear equations the
  minimal $\ell^1$-norm near solution approximates the sparest solution,''
  \emph{Communications on Pure and Applied Mathematics}, vol.~59, no.~7, pp.
  907--934, 2006.

\bibitem{BrucksteinA2007}
A.~Bruckstein, D.~Donoho, and M.~Elad, ``From sparse solutions of systems of
  equations to sparse modeling of signals and images,'' \emph{SIAM Review},
  vol.~51, no.~1, pp. 34--81, 2009.

\bibitem{ChenS2001-SIAM}
S.~Chen, D.~Donoho, and M.~Saunders, ``Atomic decomposition by basis pursuit,''
  \emph{SIAM Review}, vol.~43, no.~1, pp. 129--159, 2001.

\bibitem{CandesE2006-CPAM}
E.~Cand{\`{e}}s, J.~Romberg, and T.~Tao, ``Stable signal recovery from
  incomplete and inaccurate measurements,'' \emph{Communications on Pure and
  Applied Math}, vol.~59, no.~8, pp. 1207--1223, 2006.

\bibitem{GemmekeHCB10}
J.~F. Gemmeke, H.~V. Hamme, B.~Cranen, and L.~Boves, ``Compressive sensing for
  missing data imputation in noise robust speech recognition,'' \emph{J. Sel.
  Topics Signal Processing}, vol.~4, no.~2, pp. 272--287, 2010.

\bibitem{YangJ2008-CVPR}
J.~Yang, J.~Wright, T.~Huang, and Y.~Ma, ``Image super-resolution as sparse
  representation of raw image patches,'' in \emph{Proceedings of the {IEEE}
  International Conference on Computer Vision and Pattern Recognition}, 2008.

\bibitem{BaronD2005}
D.~Baron, M.~Wakin, M.~Duarte, S.~Sarvotham, and R.~Baraniuk, ``Distributed
  compressed sensing,'' \emph{preprint}, 2005.

\bibitem{YangA2010-PIEEE}
A.~Yang, M.~Gastpar, R.~Bajcsy, and S.~Sastry, ``Distributed sensor perception
  via sparse representation,'' \emph{Proceedings of the IEEE}, vol.~98, no.~6,
  pp. 1077--1088, 2010.

\bibitem{WrightJ2010-PIEEE}
J.~Wright, Y.~Ma, J.~Mairal, G.~Sapiro, T.~Huang, and S.~Yan, ``Sparse
  representation for computer vision and pattern recognition,''
  \emph{Proceedings of the IEEE}, vol.~98, no.~6, pp. 1031--1044, 2010.

\bibitem{WrightJ2009-PAMI}
J.~Wright, A.~Yang, A.~Ganesh, S.~Sastry, and Y.~Ma, ``Robust face recognition
  via sparse representation,'' \emph{{IEEE} Transactions on Pattern Analysis
  and Machine Intelligence}, vol.~31, no.~2, pp. 210 -- 227, 2009.

\bibitem{WagnerA2009-CVPR}
A.~Wagner, J.~Wright, A.~Ganesh, Z.~Zhou, and Y.~Ma, ``Toward a practical face
  recognition: Robust pose and illumination via sparse representation,'' in
  \emph{Proceedings of the {IEEE} International Conference on Computer Vision
  and Pattern Recognition}, 2009.

\bibitem{LorisI2009}
I.~Loris, ``On the performance of algorithms for the minimization of
  $\ell_1$-penalized functionals,'' \emph{Inverse Problems}, vol.~25, pp.
  1--16, 2009.

\bibitem{BeckerS2009}
S.~Becker, J.~Bobin, and E.~Candes, ``{NESTA}: a fast and accurate first-order
  method for sparse recovery,'' \emph{preprint}, 2009.

\bibitem{Zibulevsky2010-SPM}
M.~Zibulevskyb and M.~Elad, ``{L}1-{L}2 optimization in signal and image
  processing,'' \emph{IEEE Signal Processing Magazine}, vol.~27, no.~3, pp. 76
  -- 88, 2010.

\bibitem{DonohoD2009-PNAS}
D.~Donoho, A.~Maleki, and A.~Montanari, ``Message-passing algorithms for
  compressed sensing,'' \emph{PNAS}, vol. 106, no.~45, pp. 18\,914--18\,919,
  2009.

\bibitem{Bertsekas1982}
D.~Bertsekas, \emph{Constrained Optimization and Lagrange Multiplier
  Methods}.\hskip 1em plus 0.5em minus 0.4em\relax Athena Scientific, 1982.

\bibitem{YangJ2009}
J.~Yang and Y.~Zhang, ``Alternating direction algorithms for
  {$\ell_1$}-problems in compressive sensing,'' \emph{(preprint)
  arXiv:0912.1185}, 2009.

\bibitem{NesterovY1983-SMD}
Y.~Nesterov, ``A method of solving a convex programming problem with
  convergence rate \textit{O}$(1/k^2)$,'' \emph{Soviet Mathematics Doklady},
  vol.~27, no.~2, pp. 372--376, 1983.

\bibitem{BeckA2009}
A.~Beck and M.~Teboulle, ``A fast iterative shrinkage-thresholding algorithm
  for linear inverse problems,'' \emph{SIAM Journal on Imaging Sciences},
  vol.~2, no.~1, pp. 183--202, 2009.

\bibitem{Elad2007-ACHA}
M.~Elad, B.~Matalon, and M.~Zibulevskyb, ``Coordinate and subspace optimization
  methods for linear least squares with non-quadratic regularization,''
  \emph{Applied and Computational Harmonic Analysis}, vol.~23, no.~3, pp. 346
  -- 367, 2007.

\bibitem{BeckerS2010}
S.~R. {Becker}, E.~J. {Cand{\`e}s}, and M.~{Grant}, ``{Templates for Convex
  Cone Problems with Applications to Sparse Signal Recovery},'' \emph{ArXiv
  e-prints}, 2010.

\bibitem{FigueiredoM2007}
M.~Figueiredo, R.~Nowak, and S.~Wright, ``Gradient projection for sparse
  reconstruction: {A}pplication to compressed sensing and other inverse
  problems,'' \emph{IEEE Journal of Selected Topics in Signal Processing},
  vol.~1, no.~4, pp. 586--597, 2007.

\bibitem{WrightS2008}
S.~Wright, R.~Nowak, and M.~Figueiredo, ``Sparse reconstruction by separable
  approximation,'' in \emph{IEEE International Conference on Acoustics, Speech
  and Signal Processing}, 2008.

\bibitem{BergFriedlander2008}
\BIBentryALTinterwordspacing
E.~van~den Berg and M.~P. Friedlander, ``Probing the pareto frontier for basis
  pursuit solutions,'' \emph{SIAM Journal on Scientific Computing}, vol.~31,
  no.~2, pp. 890--912, 2008. [Online]. Available:
  \url{http://link.aip.org/link/?SCE/31/890}
\BIBentrySTDinterwordspacing

\bibitem{Afonso2010-TIP}
M.~V. Afonso, J.~M. Bioucas-Dias, and M.~A.~T. Figueiredo, ``Fast image
  recovery using variable splitting and constrained optimization,'' \emph{IEEE
  Transactions on Image Processing}, vol.~19, no.~9, pp. 2345 -- 2356, 2010.

\bibitem{FriedmanJ2010}
J.~Friedman, T.~Hastie, and R.~Tibshirani, ``Regularization paths for
  generalized linear models via coordinate descent,'' \emph{Journal of
  Statistical Software}, vol.~33, no.~1, pp. 1--22, 2010.

\bibitem{YinW2008}
W.~Yin, S.~Osher, D.~Goldfarb, and J.~Darbon, ``Bregman iterative algorithms
  for $\ell_1$-minimization with applications to compressed sensing,''
  \emph{SIAM Journal on Imaging Sciences}, vol.~1, no.~1, pp. 143--168, 2008.

\bibitem{DavisG1997}
G.~Davis, S.~Mallat, and M.~Avellaneda, ``Adaptive greedy approximations,''
  \emph{Journal of Constructive Approximation}, vol.~13, pp. 57--98, 1997.

\bibitem{NeedellD2008}
D.~Needell and J.~Tropp, ``{CoSaMP}: Iterative signal recovery from incomplete
  and inaccurate samples,'' \emph{Appl. Comp. Harmonic Anal.}, vol.~26, pp.
  301--321, 2008.

\bibitem{DaiW2009}
W.~Dai and O.~Milenkovic, ``Subspace pursuit for compressive sensing signal
  reconstruction,'' \emph{{IEEE} Transactions on Information Theory}, vol.~55,
  no.~5, pp. 2230--2249, 2009.

\bibitem{TroppJ2004}
J.~Tropp, ``Greed is good: Algorithmic results for sparse approximation,''
  \emph{{IEEE} Transactions on Information Theory}, vol.~50, no.~10, pp.
  2231--2242, 2004.

\bibitem{TroppJ2010}
J.~Tropp and S.~Wright, ``Computational methods for sparse solution of linear
  inverse problems,'' \emph{Proceedings of the IEEE}, vol.~98, pp. 948--958,
  2010.

\bibitem{ZhangL2011-ICCV}
L.~Zhang, M.~Yang, and X.~Feng, ``Sparse representation or collaborative
  representation: Which helps face recognition?'' in \emph{Proceedings of the
  {IEEE} International Conference on Computer Vision}, 2011.

\bibitem{WrightJ2011-arXiv}
J.~Wright, A.~Ganesh, A.~Yang, Z.~Zhou, and Y.~Ma, ``Sparsity and robustness in
  face recognition -- a tutorial on how to apply the models and tools
  correctly,'' in \emph{arXiv:1111.1014v1}, 2011.

\bibitem{BelhumeurP1997-PAMI}
P.~Belhumeur, J.~Hespanda, and D.~Kriegman, ``{E}igenfaces vs. {F}isherfaces:
  recognition using class specific linear projection,'' \emph{{IEEE}
  Transactions on Pattern Analysis and Machine Intelligence}, vol.~19, no.~7,
  pp. 711--720, 1997.

\bibitem{BasriR2003-PAMI}
R.~Basri and D.~Jacobs, ``Lambertian reflectance and linear subspaces,''
  \emph{{IEEE} Transactions on Pattern Analysis and Machine Intelligence},
  vol.~25, no.~2, pp. 218--233, 2003.

\bibitem{WrightJ2008-IT}
J.~Wright and Y.~Ma, ``Dense error correction via {$\ell^1$}-minimization,''
  \emph{{IEEE} Transactions on Information Theory}, vol.~56, no.~7, pp.
  3540--3560, 2010.

\bibitem{FrischK1955}
K.~Frisch, ``The logarithmic potential method of convex programming,''
  University Institute of Economics (Oslo, Norway), Tech. Rep., 1955.

\bibitem{KarmarkarN1984}
N.~Karmarkar, ``A new polynomial time algorithm for linear programming,''
  \emph{Combinatorica}, vol.~4, pp. 373--395, 1984.

\bibitem{MegiddoN1989}
N.~Megiddo, ``Pathways to the optimal set in linear programming,'' in
  \emph{Progress in Mathematical Programming: Interior-Point and Related
  Methods}, 1989, pp. 131--158.

\bibitem{MonteiroR1989-I}
R.~Monteiro and I.~Adler, ``Interior path following primal-dual algorithms.
  {P}art {I}: Linear programming,'' \emph{Mathematical Programming}, vol.~44,
  pp. 27--41, 1989.

\bibitem{KojimaM1993}
M.~Kojima, N.~Megiddo, and S.~Mizuno, ``Theoretical convergence of large-step
  primal-dual interior point algorithms for linear programming,''
  \emph{Mathematical Programming}, vol.~59, pp. 1--21, 1993.

\bibitem{BoydS2004}
S.~Boyd and L.~Vandenberghe, \emph{Convex optimization}.\hskip 1em plus 0.5em
  minus 0.4em\relax Cambridge University Press, 2004.

\bibitem{KimS2007}
S.~Kim, K.~Koh, M.~Lustig, S.~Boyd, and D.~Gorinevsky, ``An interior-point
  method for large-scale $\ell_1$-regularized least squares,'' \emph{IEEE
  Journal of Selected Topics in Signal Processing}, vol.~1, no.~4, pp.
  606--617, 2007.

\bibitem{KelleyC1995}
C.~Kelley, \emph{Iterative methods for linear and nonlinear equations}.\hskip
  1em plus 0.5em minus 0.4em\relax Philadelphia: SIAM, 1995.

\bibitem{NocedalJ2006}
J.~Nocedal and S.~Wright, \emph{Numerical Optimization}, 2nd~ed.\hskip 1em plus
  0.5em minus 0.4em\relax New York: Springer, 2006.

\bibitem{OsborneM2000}
M.~Osborne, B.~Presnell, and B.~Turlach, ``A new approach to variable selection
  in least squares problems,'' \emph{IMA Journal of Numerical Analysis},
  vol.~20, pp. 389--404, 2000.

\bibitem{EfronB2004}
B.~Efron, T.~Hastie, I.~Johnstone, and R.~Tibshirani, ``Least angle
  regression,'' \emph{The Annals of Statistics}, vol.~32, no.~2, pp. 407--499,
  2004.

\bibitem{MalioutovD2005}
D.~Malioutov, M.~Cetin, and A.~Willsky, ``Homotopy continuation for sparse
  signal representation,'' in \emph{Proceedings of the {IEEE} International
  Conference on Acoustics, Speech, and Signal Processing}, 2005.

\bibitem{DonohoD2006}
D.~Donoho and Y.~Tsaig, ``Fast solution of $\ell^1$-norm minimization problems
  when the solution may be sparse,'' \emph{{\rm preprint},
  http://www.stanford.edu/~tsaig/research.html}, 2006.

\bibitem{AsifM2008}
M.~Asif, ``Primal dual prusuit: A homotopy based algorithm for the dantzig
  selector,'' {M. S. Thesis}, Georgia Institute of Technology, 2008.

\bibitem{CombettesP2005}
P.~Combettes and V.~Wajs, ``Signal recovery by proximal forward-backward
  splitting,'' \emph{SIAM Multiscale Modeling and Simulation}, vol.~4, pp.
  1168--1200, 2005.

\bibitem{DaubechiesI2004}
I.~Daubechies, M.~Defrise, and C.~Mol, ``An iterative thresholding algorithm
  for linear inverse problems with a sparsity constraint,''
  \emph{Communications on Pure and Applied Math}, vol.~57, pp. 1413--1457,
  2004.

\bibitem{HaleE2007}
E.~Hale, W.~Yin, and Y.~Zhang, ``A fixed-point continuation method for
  $\ell^1$-regularized minimization with applications to compressed sensing,''
  Rice University, Houston, TX, Tech. Rep. CAAM Technical Report TR07-07, 2007.

\bibitem{BarzilaiJ1988}
J.~Barzilai and J.~Borwein, ``Two point step size gradient methods,'' \emph{IMA
  Journal of Numerical Analysis}, vol.~8, pp. 141--148, 1988.

\bibitem{HagerPZ11}
W.~W. Hager, D.~T. Phan, and H.~Zhang, ``Gradient-based methods for sparse
  recovery,'' \emph{SIAM J. Imaging Sciences}, vol.~4, no.~1, pp. 146--165,
  2011.

\bibitem{NesterovY2007}
Y.~Nesterov, ``Gradient methods for minimizing composite objective function,''
  \emph{ECORE Discussion Paper}, 2007.

\bibitem{BaronD2010}
D.~Baron, S.~Sarvotham, and R.~Baraniuk, ``Bayesian compressive sensing via
  belief propagation,'' \emph{{IEEE} Transactions on Signal Processing},
  vol.~58, no.~1, pp. 269--280, 2010.

\bibitem{KschischangF2001-TIT}
F.~Kschischang, B.~Frey, and H.~Loeliger, ``Factor graphs and the sum-product
  algorithm,'' \emph{{IEEE} Transactions on Information Theory}, vol.~47,
  no.~2, pp. 498--519, 2001.

\bibitem{Tseng2008}
P.~Tseng, ``On accelerated proximal gradient methods for convex-concave
  optimization,'' \emph{preprint}, 2008.

\bibitem{AuslenderJ2006-SJO}
A.~Auslender and M.~Teboulle, ``Interior gradient and proximal methods for
  convex and conic optimization,'' \emph{{SIAM} Journal on Optimization},
  vol.~16, no.~3, pp. 697--725, 2006.

\bibitem{BertsekasD2003}
D.~P. Bertsekas, \emph{Nonlinear Programming}.\hskip 1em plus 0.5em minus
  0.4em\relax Athena Scientific, 2003.

\bibitem{GoldsteinO09}
T.~Goldstein and S.~Osher, ``The split bregman method for l1-regularized
  problems,'' \emph{SIAM J. Imaging Sciences}, vol.~2, no.~2, pp. 323--343,
  2009.

\bibitem{Afonso2011-TIP}
M.~V. Afonso, J.~M. Bioucas-Dias, and M.~A.~T. Figueiredo, ``An augmented
  lagrangian approach to the constrained optimization formulation of imaging
  inverse problems,'' \emph{IEEE Transactions on Image Processing}, vol.~20,
  no.~3, pp. 681 -- 695, 2011.

\bibitem{GrossR2006}
R.~Gross, I.~Mathews, J.~Cohn, T.~Kanade, and S.~Baker, ``{Multi-PIE},'' in
  \emph{IEEE International Conference on Automatic Face and Gesture
  Recognition}, 2008.

\bibitem{ShiaV2011-Asilomar}
V.~Shia, A.~Yang, and S.~Sastry, ``Fast $\ell_1$-minimization and algorithm
  parallelization for face recognition,'' in \emph{Asilomar Conference on
  Signals, Systems and Computers}, 2011.

\end{thebibliography}
\end{document}